\pdfoutput=1
\documentclass{article}
\usepackage{preprint_style,times}

\usepackage{amsmath}
\usepackage{amssymb}
\usepackage{amsthm}
\usepackage{booktabs}
\usepackage{multirow}
\usepackage{array}
\usepackage{graphicx}
\usepackage{xcolor}
\usepackage{pifont}
\usepackage[most]{tcolorbox}
\usepackage{listings}
\usepackage{algorithm}
\usepackage{algorithmic}
\usepackage{titletoc}
\usepackage{placeins}
\usepackage{microtype}
\usepackage{hyperref}
\usepackage{url}

\hypersetup{pdftitle={Hesitation Has a Geometry: Entropy-Trained Hyperbolic Probes for Sparse
            Activation Steering},
            pdfauthor={Zeyong Zhang, Tung Sum Thomas Kwok, Tengfei Ma, Mengjia Xu}}

\newcommand{\Dball}{\mathbb{D}^{n}_{c}}
\newcommand{\Mob}{\oplus_{c}}
\newcommand{\gradR}{\nabla^{\mathcal{R}}}
\newcommand{\HEST}{\textsc{HEST}}

\newtheorem{proposition}{Proposition}

\definecolor{hestblue}{HTML}{2A78D6}
\definecolor{forkorange}{HTML}{EB6834}
\newcommand{\cmark}{\textcolor[HTML]{1A7F4B}{\ding{51}}}
\newcommand{\xmark}{\textcolor[HTML]{C0392B}{\ding{55}}}

\tcbset{outputbox/.style={enhanced, breakable, listing only, boxrule=0.6pt, arc=2pt,
        left=4pt, right=4pt, top=2pt, bottom=2pt, fonttitle=\small\bfseries,
        coltitle=white, before skip=6pt, after skip=6pt}}
\newtcbinputlisting{\outputfile}[3]{outputbox, listing file={#3}, colback=#1!4,
        colframe=#1!80!black, colbacktitle=#1!80!black, title={#2}}
\newtcblisting{promptbox}[1]{outputbox, colback=gray!4, colframe=gray!70!black,
        colbacktitle=gray!70!black, title={#1}}

\title{Hesitation Has a Geometry: \\
Entropy-Trained Hyperbolic Probes for \\
Sparse Activation Steering}

\author{\makebox[0.47\textwidth][l]{Zeyong Zhang} \\
Department of Data Science \\
New Jersey Institute of Technology \\
Newark, NJ, USA \\
\texttt{zz63@njit.edu}
\And
\makebox[0.47\textwidth][l]{Tung Sum Thomas Kwok} \\
Department of Statistics and Data Science \\
University of California, Los Angeles \\
Los Angeles, CA, USA \\
\texttt{tk1018@ucla.edu}
\AND
\makebox[0.47\textwidth][l]{Tengfei Ma} \\
Department of Biomedical Informatics \\
Stony Brook University \\
Stony Brook, NY, USA \\
\texttt{tengfei.ma@stonybrook.edu}
\And
\makebox[0.47\textwidth][l]{Mengjia Xu\thanks{Corresponding author.}} \\
Department of Data Science \\
New Jersey Institute of Technology \\
Newark, NJ, USA \\
\texttt{mx6@njit.edu}}

\begin{document}

\maketitle

\begin{abstract}

When a large language model solves a mathematical problem, its reasoning is largely hierarchical, and the solution often branches at a few tokens where the next-token entropy is high. Such tree-like structure embeds in hyperbolic space with far lower distortion than in Euclidean space. Activation steering, however, usually edits the hidden states of a pretrained model by adding one fixed Euclidean vector at every token, even though most tokens of a solution are already determined by the context. We propose \textbf{H}yperbolic \textbf{E}ntropy \textbf{St}eering (\HEST{}), which embeds the hidden states in the Poincar\'e ball with a lightweight probe whose only label is the model's own next-token entropy. Where this entropy exceeds a threshold, \HEST{} moves the embedded state along the geodesic of steepest descent of a readout of the probe and maps the change back to the hidden state. For the Busemann readout of a learned ideal point, we prove that a step of fixed length lowers it by the same amount at every state. On three instruction-tuned models from the Qwen2.5-Math and Llama-3.1 families, with the step length selected on held-out training problems, \HEST{} with the Busemann readout improves the accuracy of greedy decoding on MATH-500 and GSM8K in all six settings, by up to $1.1$ points, while steering a median of only $0.8$ to $3.0\%$ of the tokens of a solution, whereas a contrastive steering vector added at every token lowers accuracy. The gains are largest on problems where the model hesitates often, and accuracy on the remaining problems is almost unchanged.

\end{abstract}

\section{Introduction}\label{sec:intro}
\begin{figure}[t]
\centering
\includegraphics[width=\textwidth]{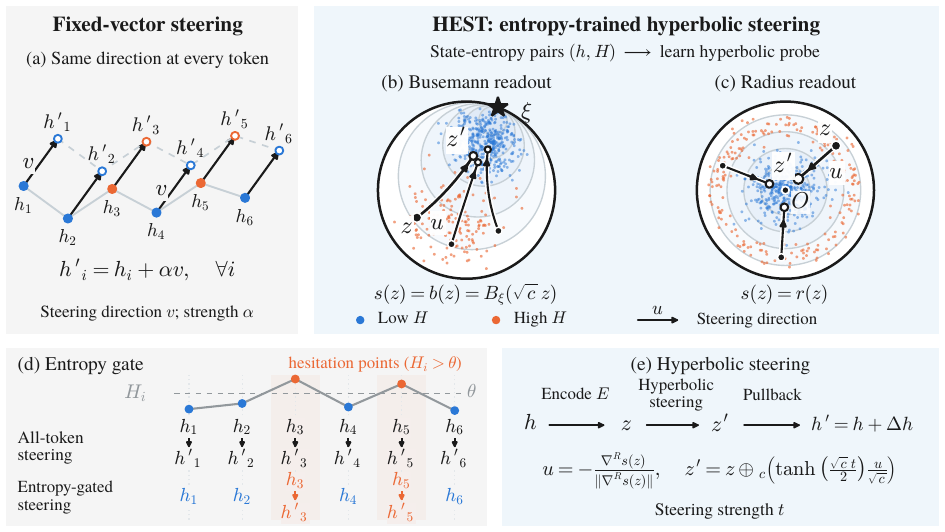}
\caption{\textbf{\HEST{} steers only the tokens at which the model hesitates, along a geodesic in
the space of a hyperbolic probe.} (a)~Fixed-vector steering adds the same vector $\alpha v$
at every token. (b,\,c)~\HEST{} maps hidden states into the Poincar\'e ball with a probe trained
on next-token entropy. The readout $s$ is either the Busemann function of the ideal point $\xi$
or the distance $r$ to the origin $O$, and the arrows show the steering direction $u$.
(d)~\HEST{} steers only tokens whose entropy $H_i$ exceeds the threshold $\theta$. (e)~At
such a token, $h$ is mapped to $z$, moved by hyperbolic distance $t$ along the geodesic in the
direction $u$, and mapped back. In (e), $\lVert\cdot\rVert$ is the Euclidean norm. The point clouds and step lengths in (b,\,c) are
schematic, and Appendix~\ref{app:disks} shows real embedded states.}
\label{fig:pipeline}
\end{figure}

Large language models solve mathematical problems through multi-step reasoning, and such reasoning, like language itself, is largely hierarchical. Hierarchical relations between concepts are reflected in the geometry of their representations \citep{park2025geometry}. In mathematical reasoning, a chain of thought also tends to branch at a small number of high-entropy tokens, which can determine the subsequent solution path \citep{wang2026beyond}. Hyperbolic space suits such tree-like structure, since its volume grows exponentially with the radius, and trees embed in it with far lower distortion than in a Euclidean space of the same dimension \citep{nickel2017poincare,sala2018representation}. This property has been used to model hierarchies in neural networks, Transformers and language models \citep{ganea2018hyperbolic,yang2024hypformer,yang2026hyperbolic,he2026helm,patil2026hyperbolic,patil2026hierarchical}, and it also makes hyperbolic space a natural candidate for representing the hidden states of Euclidean language models.

Activation steering changes the behavior of a pretrained language model at inference time without updating its weights.
It shifts the hidden activations along a direction extracted from contrastive examples \citep{turner2023steering,li2023inference,zou2023representation,rimsky2024steering}. In activation-addition methods, this direction is a Euclidean vector that is typically the same at every token position (Figure~\ref{fig:pipeline}a). Contrastive Activation Addition \citep[CAA;][]{rimsky2024steering}, for example, applies the same steering vector throughout generation.
In mathematical reasoning, such a vector perturbs every token of a solution, although most tokens are largely determined by the preceding context and the solution branches at only a few high-entropy tokens. Existing steering methods for mathematical reasoning mostly act at the step boundaries of long reasoning models \citep{chen2025seal,li2026steering} and classify the steps by keywords such as ``Wait'', which instruction-tuned models rarely produce. We instead steer instruction-tuned language models only at high-entropy tokens, and we take the direction from a hyperbolic embedding of their hidden states that a lightweight probe learns from the same entropy.

We start from an observation about the model's own uncertainty. The entropy of the next-token
distribution is low at most positions and exceeds a threshold at only a small fraction of them,
which we call \emph{hesitation points}. The same entropy can also serve as a training signal, and we
train a probe that maps hidden states into the Poincar\'e ball and orders them by entropy. Without
external labels, it learns a geometry in which confident states lie closer than hesitant ones to an
ideal point on the boundary of the ball, or to its origin. The Busemann function of this ideal point and the distance to the origin serve as
readouts, and the negative Riemannian gradient of a readout gives a steering direction that depends
on the state. We build Hyperbolic Entropy
Steering (\HEST{}) on this observation (Figure~\ref{fig:pipeline}). During greedy decoding, \HEST{}
embeds the hidden state at every hesitation point and moves the embedding along a geodesic that
lowers the readout. It then maps the change
back to the hidden state, decodes the next token from the edited state and edits no other
position. We show that a step of hyperbolic length $t$ lowers the Busemann readout by
exactly $\sqrt{c}\,t$, where $-c$ is the curvature, and the radius readout by exactly $t$ whenever
the step does not pass the origin. Hence, the
step length has the same meaning at every state (Proposition~\ref{prop:step}).

We evaluate \HEST{} on Qwen2.5-Math-1.5B-Instruct, our development model,
Qwen2.5-Math-7B-Instruct and Llama-3.1-8B-Instruct, using MATH-500 and GSM8K. The step length
is selected on held-out training problems, and every setting is run with three probes trained
with different random seeds. With the Busemann readout, \HEST{} improves the mean accuracy of greedy
decoding in all six model and dataset pairs, by $0.08$ to $1.07$ points, while steering a median
of $0.8$ to $3.0\%$ of solution tokens. On the full MATH test set of $5{,}000$ problems, it gains
$0.96$ points on the development model, with $134$ problems fixed and $86$ broken. A
difference-of-means vector built from the same states and added at every token, as in CAA, lowers
accuracy in all six pairs. With the gate and the edit size of \HEST{}, a fixed direction, a
reversed step, random directions and a Euclidean probe trained in the same way all leave the
accuracy averaged over the six pairs at or below that of greedy decoding. The gains are largest on the problems on which the greedy solution hesitates often. Most other greedy solutions are already correct, and \HEST{} rarely changes their outcome. Our contributions are:
\begin{itemize}
\item We propose \HEST{} (Section~\ref{sec:method}), an inference-time steering method in which the model's own next-token
entropy decides where to steer and also trains the hyperbolic probe that gives the direction.
The probe is trained without correctness labels or contrastive prompts, and \HEST{} needs no
language model training or additional sampled solutions.
\item We give a geometric account of the method (Sections~\ref{sec:probe} and~\ref{sec:step}, Appendix~\ref{app:geometry}). Both readouts have a direct interpretation, one
geodesic step lowers a readout by an exact amount, and in the probe space the Busemann direction
depends on the state, whereas its Euclidean counterpart does not.
\item We evaluate \HEST{} on three models against greedy decoding, CAA and HARP, confirm the gain
on the full MATH test set (Section~\ref{sec:results}), compare the probe direction with fixed, reverse, random and Euclidean
directions in every setting (Section~\ref{sec:ablation}), and analyze which problems benefit (Section~\ref{sec:where}).
\end{itemize}

\section{Related Work}\label{sec:related-work}
\paragraph{Activation steering.}
Most activation steering methods add a fixed vector to the hidden states, at every generated token
or at fixed positions of the prompt. The vector is obtained by contrasting prompts or examples on
behaviors such as truthfulness, refusal and sentiment
\citep{turner2023steering,li2023inference,zou2023representation,rimsky2024steering}. ReFT and
HyperSteer instead learn the intervention, as low-rank edits at fixed prompt positions and as
vectors that a hypernetwork generates from a steering prompt and the input \citep{wu2024reft,sun2025hypersteer}.
Several recent methods let the edit depend on the state. AcT transports activations with
coordinate-wise affine maps from optimal transport \citep{rodriguez2025controlling}, and ODESteer
and FLAS move them along a vector field \citep{zhao2026odesteer,jin2026beyond}. Curveball Steering
works in the feature space of a kernel PCA \citep{raval2026curveball}, and Angular and Spherical
Steering rotate activations \citep{vu2026angular,you2026spherical}. K-Steering, the closest to our
mechanism, follows the gradient of a nonlinear classifier trained with attribute labels
\citep{oozeer-etal-2025-beyond}. CLAS scales a direction at every layer by a coefficient linear in
the activation, and it uses GSM8K to elicit step-by-step solutions \citep{hsu2026contextual}.
Other methods restrict where the edit is applied. MAT-Steer learns a gate that weights each
steering vector at each prompt token \citep{nguyen2025multi}, and Spherical Steering rotates only
when a confidence score for the undesired direction exceeds a threshold \citep{you2026spherical}.
FASB backtracks when a probe trained on labeled data detects a deviation \citep{cheng2026steering}.
Helix lowers the largest logit where a score of the next-token entropy and a Mahalanobis distance
of the hidden state indicates divergence \citep{atkinson2026tethered}. \HEST{} combines a
state-dependent direction with a sparse gate, and both come from the model's own entropy rather
than from labels, as the comparison in Appendix~\ref{app:position} details.

\paragraph{Entropy and forking tokens.}
Token-level entropy marks the positions where a reasoning chain often branches. In
reinforcement learning for reasoning, a minority of high-entropy tokens determines the
reasoning path, and restricting policy updates to these tokens matches or exceeds updating all
tokens \citep{wang2026beyond}. Hidden states also reflect uncertainty about which path the
generation will take \citep{zur2025language}. HARP treats tokens whose entropy
exceeds a threshold as points of hesitation, applies dropout to the input embeddings and mixes the
logits of a second forward pass \citep{storai2025harp}. Entropy-gated branching instead selects
among several continuations at such tokens with a process reward model
\citep{li2026entropy}. Some steering methods for mathematical reasoning
target long reasoning models. SEAL splits the reasoning into steps at double line breaks, classifies the
steps by keywords and steers the representations at every step boundary \citep{chen2025seal}.
FlowSteer keeps this protocol but replaces the linear edit with flow matching
\citep{li2026steering}. \citet{sun2026llm} study the reasoning trajectories of three
Llama-3.1-8B variants and also steer at step boundaries. Their steering uses a predictor trained
on correctness labels, or ideal trajectories from correct solutions. Like HARP and
entropy-gated branching, \HEST{} selects tokens by their entropy, but it edits their hidden states
and uses neither keywords nor correctness labels.

\paragraph{Latent-space methods for test-time scaling.}
Another line of work spends additional inference compute in latent space. Fractional Reasoning
scales a latent steering vector to control the depth of reasoning, and it reports its gains with
majority voting or best-of-$N$ selection \citep{liu2025fractional}. LatentSeek optimizes latent
representations for each problem with a policy gradient at test time \citep{li2025seek}. SPREAD
jointly optimizes steering vectors for several sampled solutions to diversify them, by Riemannian
optimization over a product of spheres \citep{khanh2026test}. These methods use several
candidates or optimize per problem, whereas \HEST{} edits a single greedy decode, so we do not use
them as baselines. Random embedding injection appends untrained random embeddings to the input to diversify
reasoning, and it is complementary to \HEST{} \citep{kim2026noise}.

\paragraph{Hyperbolic representation.}
\citet{ganea2018hyperbolic} define
hyperbolic neural network layers through M\"obius gyrovector operations in the Poincar\'e ball.
\citet{ghadimi2021hyperbolic} use Busemann functions to classify against ideal prototypes, and
\citet{chen2026hbnn} build hyperbolic network layers from them. HypFormer builds fully hyperbolic
Transformers \citep{yang2024hypformer}. HypLoRA fine-tunes large language models
with low-rank adapters in hyperbolic space \citep{yang2026hyperbolic}, and HELM builds hyperbolic
large language models, including a variant whose experts have different curvatures
\citep{he2026helm}. \citet{patil2026hyperbolic} survey hyperbolic language models. HyperGuide projects
hidden states into the Poincar\'e ball and fine-tunes the model to use the resulting signal in
multi-step reasoning \citep{liu2026hyperguide}. Our probe follows the ranking and triplet
objective of HyperGuide, which trains its projection on distances to a solution in a search tree
rather than on the model's own entropy. We use our readouts to steer a
Euclidean model without fine-tuning it.

\section{Method}\label{sec:method}

\paragraph{Setup and notation.}
Let $M$ be a decoder-only language model, decoded greedily, with residual stream
$h\in\mathbb{R}^{d}$ at layer $\ell$. After the model reads the token at position $i$, it has a
layer-$\ell$ state $h_i$ and a distribution $p_i$ over the next token, with entropy
\begin{equation}
H_i \;=\; -\sum_{v\in\mathcal{V}} p_i(v)\log p_i(v).
\label{eq:entropy}
\end{equation}
It is computed from logits that the model produces during decoding, so it requires no annotation.
We call position $i$
a \emph{hesitation point} when $H_i>\theta$. The entropy plays two roles in \HEST{}: the probe
learns to order states by it (Section~\ref{sec:probe}), and it decides where to steer
(Section~\ref{sec:gate}).

\subsection{A Hyperbolic Probe Trained on Entropy}
\label{sec:probe}

We use the Poincar\'e ball $\Dball=\{z\in\mathbb{R}^{n}:c\lVert z\rVert^{2}<1\}$ of curvature
$-c$, with metric $g_z=(\lambda^{c}_{z})^{2}g^{E}$ and conformal factor
$\lambda^{c}_{z}=2/(1-c\lVert z\rVert^{2})$ \citep{ganea2018hyperbolic}. The probe
$E:\mathbb{R}^{d}\to\Dball$ is
\begin{equation}
E(h)=\exp^{c}_{0}\!\left(\frac{\phi\big((h-\mu)/\sigma\big)}{\omega}\right),
\qquad
\exp^{c}_{0}(v)=\tanh\!\big(\sqrt{c}\,\lVert v\rVert\big)\,\frac{v}{\sqrt{c}\,\lVert v\rVert},
\label{eq:probe}
\end{equation}
where $\mu$ and $\sigma$ are the coordinate-wise mean and standard deviation of the training
states, and $\phi$ is a two-layer perceptron with hidden width $1024$ and output dimension
$n=128$. The scalar $\omega$ is a running root mean square of $\lVert\phi(\cdot)\rVert$ and fixes
the scale of the embedding.

\paragraph{Readouts.}
A readout is a scalar function $s$ on the ball, low at confident states and high at
hesitant ones. The first readout is built from the Busemann function of a learned ideal point on
the boundary of the ball \citep{ghadimi2021hyperbolic}. We write this point as $\xi/\sqrt{c}$ with
$\lVert\xi\rVert=1$ and define
\begin{equation}
b(z)=B_{\xi}\big(\sqrt{c}\,z\big),\qquad
B_{\xi}(y)=\log\frac{\lVert\xi-y\rVert^{2}}{1-\lVert y\rVert^{2}} .
\label{eq:busemann}
\end{equation}
Here $B_{\xi}$ is the Busemann function of the unit ball, so $b$ is $\sqrt{c}$ times that of $\Dball$. Its level sets are horospheres, which are spheres tangent to the boundary at the ideal point,
and $b$ decreases without bound along every geodesic that ends at $\xi/\sqrt{c}$. The second readout is the distance to the
origin,
\begin{equation}
r(z)=d_{c}(0,z)=\frac{2}{\sqrt{c}}\operatorname{artanh}\!\big(\sqrt{c}\,\lVert z\rVert\big),
\label{eq:radius}
\end{equation}
whose level sets are spheres centered at the origin. Both readouts are interpretable, as confident
states tend to lie on smaller horospheres than hesitant ones, or nearer to the
origin (Figure~\ref{fig:pipeline}b,c).

\paragraph{Training objective.}
Write $z_i=E(h_i)$. We follow \citet{liu2026hyperguide}, who train a hyperbolic probe with a
pairwise margin ranking loss \citep{herbrich2000large} and a triplet loss
\citep{weinberger2009distance,schroff2015facenet} in the Poincar\'e ball. The probe minimizes
\begin{equation}
\mathcal{L}=
\mathop{\mathbb{E}}_{(i,j):\,H_i<H_j-\delta}\Big[\gamma+s(z_i)-s(z_j)\Big]_{+}
+\mathop{\mathbb{E}}_{(a,p,q)}\Big[\gamma+d_{c}(z_a,z_p)-d_{c}(z_a,z_q)\Big]_{+},
\label{eq:loss}
\end{equation}
with $[x]_{+}=\max(0,x)$, $\delta=0.3$ nats and $\gamma=0.1$. The first term orders states by
entropy, and the second keeps nearby states of one solution close: the positive $p$ lies within
five tokens of the anchor $a$, and the negative $q$ comes from another
problem. The curvature and the ideal point are learned jointly with $\phi$. The data are the
states and entropies of greedy solutions to $100$ training problems, without any correctness
label. On held-out problems, the Busemann readout orders $85$ to $90\%$ of the entropy pairs
correctly in all settings, and the radius readout $82$ to $89\%$ (Appendix~\ref{app:training}).

\subsection{Steering along a Geodesic}
\label{sec:step}

Let $h$ be the state at a hesitation point and $z=E(h)$. Because the metric is conformal,
the Riemannian gradient $\gradR s(z)=(\lambda^{c}_{z})^{-2}\,\nabla s(z)$ has the direction of
the Euclidean gradient. We set
\begin{equation}
u=-\frac{\gradR s(z)}{\lVert\gradR s(z)\rVert},
\qquad
z'=z\Mob\left(\tanh\!\Big(\frac{\sqrt{c}\,t}{2}\Big)\frac{u}{\sqrt{c}}\right),
\label{eq:geostep}
\end{equation}
where $\lVert\cdot\rVert$ is the Euclidean norm and $\Mob$ is M\"obius addition
\citep{ganea2018hyperbolic}. The point $z'$ equals $\exp^{c}_{z}(t\,v)$ with $v=u/\lambda^{c}_{z}$, the unit
tangent vector in the direction $u$ under the hyperbolic metric.

\begin{proposition}
\label{prop:step}
Let $z'$ be given by Eq.~\ref{eq:geostep} with $t>0$ and $\nabla s(z)\neq0$. Then
(i) $d_{c}(z,z')=t$;
(ii) for the radius readout, if $z\neq 0$ and $t\leq r(z)$, then $r(z')=r(z)-t$;
(iii) for the Busemann readout, $b(z')=b(z)-\sqrt{c}\,t$.
\end{proposition}

Part (i) follows from the left-cancellation law of M\"obius addition. Parts (ii) and (iii)
use two standard facts: distance and Busemann functions have unit gradient in the
hyperbolic metric, and their descent curves are geodesics \citep{bridson2013metric}. The factor $\sqrt{c}$ comes
from the rescaling $y=\sqrt{c}\,z$, and Appendix~\ref{app:geometry} gives the proof. The step length
$t$ thus has the same meaning at every state, since it is the hyperbolic distance traveled and
fixes the drop of the readout. Proposition~\ref{prop:step} is a statement about the readout of the
probe, and Appendix~\ref{app:mechanism} reports how the edit changes the next-token distribution of the model. The condition $t\le r(z)$ holds at every
hesitation point of all greedy test solutions (Appendix~\ref{app:radius}).

\paragraph{Role of the curvature.}
In the limit $c\to0$, evaluated in Section~\ref{sec:ablation}, the probe becomes Euclidean, the
geodesic step becomes a straight step $z'=z+(t/2)\,u$, and the readouts become the norm and the
linear function $-\langle\xi,z\rangle$ up to scale (Appendix~\ref{app:euclid}). For the radius
readout, the step points to the origin in both geometries, and the two differ only in how far a
step of length $t$ moves. The Busemann readout
is where the geometries differ. In the Poincar\'e ball, its descent curves are geodesics that
converge on the ideal point, so the direction $u$ depends on the state. The linear
function, in contrast, has parallel hyperplanes as level sets and the same descent direction
$\xi$ everywhere (Appendix~\ref{app:direction}). After the mapping back to the hidden state, any
readout gives a state-dependent direction in $\mathbb{R}^{d}$ through the Jacobian of the probe (Appendix~\ref{app:field}), so
the geometries differ in the probe space.

\subsection{Mapping the Step back to the Hidden State}
\label{sec:pullback}

We look for an edit $\Delta h$ with $E(h+\Delta h)\approx z'$ and solve this underdetermined
problem by damped Gauss--Newton iterations from $\Delta h=0$,
\begin{equation}
\Delta h \;\leftarrow\; \Delta h + J^{\top}\big(JJ^{\top}+\eta I\big)^{-1}
\big(z'-E(h+\Delta h)\big),
\qquad J=\frac{\partial E}{\partial h}\Big|_{h+\Delta h},
\label{eq:pullback}
\end{equation}
with $\eta=10^{-3}\operatorname{tr}(JJ^{\top})/n$. We run at most four iterations and clip
$\lVert\Delta h\rVert$ to at most $\lVert h\rVert$. Each iteration solves an $n\times n$
system, and $J$ has a closed form.
The edited states realize the distance $t$ and the drop $\sqrt{c}\,t$ of the Busemann readout in
Proposition~\ref{prop:step} to within $10^{-4}$ on average (Appendix~\ref{app:jacobian}).

\subsection{Decoding with the Entropy Gate}
\label{sec:gate}

\HEST{} runs inside greedy decoding, where the model produces $p_i$ after reading the token $y_i$.
If $H_i>\theta$, we remove the key and value entries of position $i$ from the cache. We then
recompute the forward pass of $y_i$ with $h_i+\Delta h_i$ in place of $h_i$ at layer $\ell$ and
choose the next token from the new distribution. The edit thus acts on the state that
produced the uncertain distribution, and one position later the model is usually confident again
(Appendix~\ref{app:decoding}). The edited keys and values stay
in the cache, so later tokens attend to the edited state. No other position is edited
(Algorithm~\ref{alg:hest}). A steered decode that reaches the limit of $3072$ new tokens falls
back to the greedy answer.

\subsection{Choice of Hyperparameters}
\label{sec:calib}

The three hyperparameters of \HEST{} are the layer $\ell$, the threshold $\theta$ and the step
length $t$, and we take the layer from prior work. For the Qwen models, we use
$\ell=19$, the layer that SEAL uses for DeepSeek-R1-Distill-Qwen-1.5B and 7B, which are fine-tuned
from Qwen2.5-Math \citep{chen2025seal}. For Llama, we use $\ell=13$, the layer that CAA uses for
Llama 2 7B Chat and ODESteer uses for Llama-3.1-8B \citep{rimsky2024steering,zhao2026odesteer},
and Appendix~\ref{app:training} explains the layer indexing. A middle layer suits \HEST{} because the probe must read the hesitation from the state, and the edit, which enters the keys and values of all later layers, needs enough layers after it (Appendix~\ref{app:training}). Appendix~\ref{app:ablation} compares
layer $19$ with layers $9$ and $25$. The threshold is $\theta=1$
for the development model, Qwen2.5-Math-1.5B-Instruct. Since $\exp(H_i)$ is the effective number
of candidate tokens, $H_i>1$ means more than $e\approx2.7$ effective candidates.
Because entropy scales differ across models, we choose $\theta$ for each other model so that
\HEST{} steers about as many tokens per solution as on the development model. We estimate this
number before steering from the teacher-forced entropies of the greedy solutions, which needs no
labels (Appendix~\ref{app:theta}). The step length is selected on held-out training problems. For every model and dataset, we draw
$500$ problems from the training split that are disjoint from the $100$ problems of the probe, run
\HEST{} with $t\in\{0.1,0.25,0.5,1\}$ and keep the step length with the highest accuracy, taking
the smaller one in a tie (Table~\ref{tab:val}, Appendix~\ref{app:val}). No test label
enters any hyperparameter. We use the Busemann readout by default, because its decrease in
Proposition~\ref{prop:step} holds at every state, whereas that of the radius readout requires
$t\le r(z)$. Appendix~\ref{app:fixedt} reports the radius readout.

\section{Empirical Results}\label{sec:results}
\subsection{Setup}
\label{sec:setup}

\paragraph{Models and benchmarks.}
We use Qwen2.5-Math-1.5B-Instruct as the development model, together with
Qwen2.5-Math-7B-Instruct \citep{yang2024qwen2} and Llama-3.1-8B-Instruct
\citep{grattafiori2024llama}. We evaluate on MATH-500 \citep{hendrycks2021measuring,lightman2024let}
and on the $1{,}319$ test problems of GSM8K \citep{cobbe2021training}. For the development
model, we also evaluate on the full MATH test set of $5{,}000$ problems, which contains MATH-500.
For each dataset, the probe is trained on $100$ problems from the training split, and the step
length is selected on $500$ other training problems (Section~\ref{sec:calib}). The test labels are
used only to report accuracy. We train three probes per setting with different random seeds and report
the mean and the standard deviation of \HEST{} over them, while the run on the full MATH test set
uses one probe. All methods decode greedily from
the same prompt with a limit of $3072$ new tokens. All outputs are graded by one protocol that extracts
the final answer and compares it with the reference by string, numeric and symbolic checks
(Appendix~\ref{app:judge}).

\paragraph{Baselines.}
We compare with greedy decoding, with CAA \citep{rimsky2024steering} and with HARP
\citep{storai2025harp}. The CAA direction is the
difference of means $v=\mathbb{E}[h\mid H\le\theta]-\mathbb{E}[h\mid H>\theta]$. It uses the same
probe-training states, layer and threshold as \HEST{}. It is added at every generated token with
multiplier $1$, as in most experiments of the original method. CAA is thus the fixed-direction, all-token
counterpart of \HEST{}, built from the same data. HARP also acts at high-entropy tokens, but it
does not edit hidden states: it applies dropout to the input embeddings in a second forward pass
and mixes the logits of the two passes. We run its official code with the default hyperparameters
and report the mean and the standard deviation of three runs.
All methods use the fallback of
Section~\ref{sec:gate} (Appendices~\ref{app:caa} and~\ref{app:harp}).
We do not include SEAL and FlowSteer, which rely on keyword-labeled steps of long reasoning
models. We also do not include the method of \citet{sun2026llm}, which needs
correctness labels, or methods that use several samples.
State-dependent methods such as K-Steering, ODESteer and CLAS learn from labels or
contrastive examples (Appendix~\ref{app:position}), so Section~\ref{sec:ablation} instead trains the
probe of \HEST{} with $c\to0$.

\begin{table}[!htb]
\caption{Selection of the step length $t$ on $500$ held-out training problems per setting. Entries
are numbers of correctly solved problems with \HEST{} (Busemann readout). The selected step
length is in bold, and ties go to the smaller one.}
\label{tab:val}
\centering
\small
\setlength{\tabcolsep}{5pt}
\begin{tabular}{llccccc}
\toprule
Model & Dataset & $t=0.1$ & $t=0.25$ & $t=0.5$ & $t=1$ & Selected $t$ \\
\midrule
\multirow{2}{*}{Qwen2.5-Math-1.5B} & MATH & 413 & \textbf{416} & 414 & 410 & 0.25 \\
 & GSM8K & 478 & 480 & 481 & \textbf{482} & 1 \\
\midrule
\multirow{2}{*}{Llama-3.1-8B} & MATH & 347 & 340 & \textbf{348} & 348 & 0.5 \\
 & GSM8K & \textbf{464} & 460 & 462 & 463 & 0.1 \\
\midrule
\multirow{2}{*}{Qwen2.5-Math-7B} & MATH & 414 & 414 & 409 & \textbf{415} & 1 \\
 & GSM8K & 479 & 481 & 480 & \textbf{482} & 1 \\
\bottomrule
\end{tabular}
\end{table}

\subsection{Main Results}
\label{sec:main}

Table~\ref{tab:val} gives the selection of the step length. \HEST{} is robust to this choice: on the $500$ validation problems
of a setting, the four step lengths differ by $3$ to $8$ problems. Table~\ref{tab:models} reports the test accuracy at the selected step
length. \HEST{} improves the mean accuracy of greedy decoding in all six model and dataset pairs.
The gain is $0.47$ to $1.07$ points on MATH-500 and $0.08$ to $0.18$ points on GSM8K, where greedy
decoding already solves $85$ to $95\%$ of the problems. The gain is stable across probes, with a standard deviation of $0.12$ to $0.42$ points, and $16$ of the $18$ individual runs are at or above greedy
decoding (Appendix~\ref{app:full}).

CAA lowers accuracy in all six settings, by $0.38$ to $6.60$ points,
although it uses the same data. It changes $97$ to $100\%$ of the
solutions (Appendix~\ref{app:caa}).
HARP is below greedy decoding in three of the six settings, and \HEST{} has the highest mean
accuracy in all six. \HEST{} is also more stable, with a standard deviation of at most $0.42$ points
against up to $1.33$ for HARP, and more conservative: in one run of Llama-3.1-8B-Instruct on
MATH-500, HARP fixes $42$ problems and breaks $48$, whereas \HEST{} fixes $17$ and breaks $11$. \HEST{} steers a median of $0.8$ to
$3.0\%$ of the tokens of a solution, and each steered token costs one extra single-token forward pass of the model, besides the
probe, its Jacobian and at most four $n\times n$ solves. \HEST{} is also cheap. Counted in token forward passes,
it needs $1$ to $14\%$ more than greedy decoding with a cache. The second pass of HARP cannot reuse the cache, and for the median
problem HARP needs $5.5$ to $30$ times the passes of greedy decoding (Appendix~\ref{app:harp}).

\begin{table}[!htb]
\caption{Accuracy (\%) on the test sets. $\theta$: gate threshold. $t$: step length selected in
Table~\ref{tab:val}. \HEST{}: mean and standard deviation over three probes. HARP: mean and standard
deviation over three runs. All methods use the fallback of Section~\ref{sec:gate}. Steered: median percentage of tokens steered per solution (\%). Bold: best accuracy per row.}
\label{tab:models}
\centering
\small
\setlength{\tabcolsep}{4pt}
\begin{tabular}{ll|cc|ccc|cc}
\toprule
Model & Dataset & $\theta$ & $t$ & Greedy & CAA & HARP & \HEST{} & Steered \\
\midrule
\multirow{2}{*}{Qwen2.5-Math-1.5B} & MATH-500 & 1.0 & 0.25 & 74.20 & 72.40 & $74.13\pm0.76$ & $\mathbf{74.67}\pm0.31$ & 1.1 \\
 & GSM8K & 1.0 & 1 & 84.84 & 84.38 & $84.94\pm0.50$ & $\mathbf{85.01}\pm0.31$ & 1.6 \\
\midrule
\multirow{2}{*}{Llama-3.1-8B} & MATH-500 & 2.0 & 0.5 & 50.80 & 44.20 & $50.27\pm1.33$ & $\mathbf{51.87}\pm0.12$ & 0.8 \\
 & GSM8K & 1.4 & 0.1 & 84.61 & 80.21 & $84.46\pm1.05$ & $\mathbf{84.69}\pm0.42$ & 3.0 \\
\midrule
\multirow{2}{*}{Qwen2.5-Math-7B} & MATH-500 & 1.0 & 1 & 84.00 & 83.20 & $84.87\pm0.50$ & $\mathbf{85.07}\pm0.42$ & 1.0 \\
 & GSM8K & 0.8 & 1 & 95.38 & 95.00 & $95.53\pm0.13$ & $\mathbf{95.55}\pm0.16$ & 2.0 \\
\bottomrule
\end{tabular}
\end{table}

\subsection{The Full MATH Test Set}
\label{sec:where}

\begin{figure}[!htb]
\centering
\includegraphics[width=\textwidth]{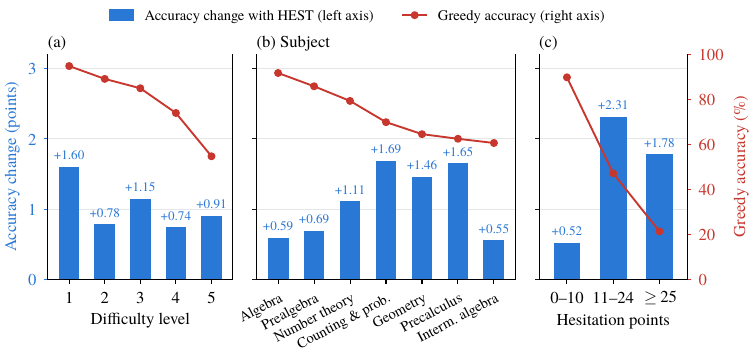}
\caption{\HEST{} improves accuracy in every part of the full MATH test set ($5{,}000$ problems,
Qwen2.5-Math-1.5B-Instruct, $t=0.25$). Bars: change in accuracy from greedy decoding. Line: accuracy
of greedy decoding. Problems are divided by (a)~difficulty level, (b)~subject and (c)~the number of
hesitation points in the greedy solution. Table~\ref{tab:app-math5000} gives the numbers.}
\label{fig:math5000}
\end{figure}

MATH-500 is a subset of the MATH test set, and a change of one point on it corresponds to five
problems. We therefore run the development model on all $5{,}000$ test problems, with the probe
and the step length $t=0.25$ of Table~\ref{tab:val} unchanged. Greedy decoding reaches $75.82\%$, while \HEST{} improves it to $76.78\%$, yielding a gain of $0.96$ points. \HEST{} fixes $134$ problems and breaks $86$, and an
exact two-sided sign test on these counts gives $p=0.0015$. Figure~\ref{fig:math5000} divides the
problems by difficulty level, subject, and the number of hesitation points in the greedy
solution. Accuracy increases in each of these $15$ parts (Appendix~\ref{app:math5000}).

\begin{figure}[!htb]
\centering
\includegraphics[width=\textwidth]{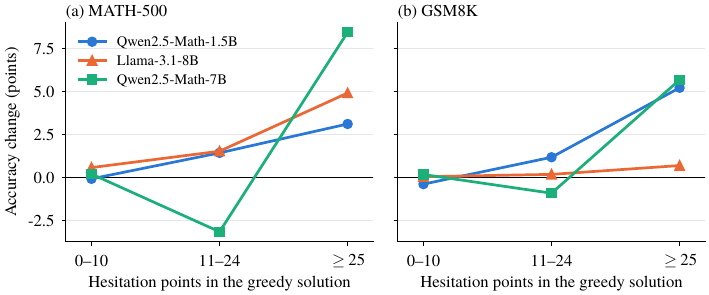}
\caption{The gain of \HEST{} is largest on problems whose greedy solution hesitates often. Change
in accuracy from greedy decoding by the number of hesitation points in the greedy solution, averaged
over the three probes, on (a)~MATH-500 and (b)~GSM8K.}
\label{fig:groups}
\end{figure}

\paragraph{Hesitation groups.}
To see which problems benefit, we divide them by the number of hesitation
points in their greedy solution, which is fixed before any intervention, into groups with $0$ to
$10$, $11$ to $24$ and at least $25$ hesitation points. On the full MATH
test set (Figure~\ref{fig:math5000}c), the first group contains $73\%$ of the problems, and greedy decoding
already solves $89.7\%$ of them. \HEST{} gains $0.52$ points there, against $2.31$ and $1.78$
points in the two groups that hesitate more, where greedy decoding solves $47.1\%$ and $21.3\%$.
In the six settings of Table~\ref{tab:models} (Figure~\ref{fig:groups}), the group with at least $25$ hesitation points has
the largest gain in every setting, $0.68$ to $8.44$ points, and the first group changes by $-0.40$
to $+0.56$ points. Summed over all settings and
probes, \HEST{} fixes $81$ problems and breaks $33$ in the group with at least $25$ hesitation points (Appendix~\ref{app:groups}).
\HEST{} thus leaves the problems that greedy decoding already solves almost unchanged
and improves those on which the model hesitates often. The reason is that a hesitation point is a close choice between candidate tokens. The edit changes the most likely token at only $2.8$ to $10.7\%$ of the steered positions, so a solution with few hesitation points is rarely changed, and a solution with many of them offers several such choices (Appendix~\ref{app:mechanism}).

\begin{figure}[!htb]
\centering
\includegraphics[width=\textwidth]{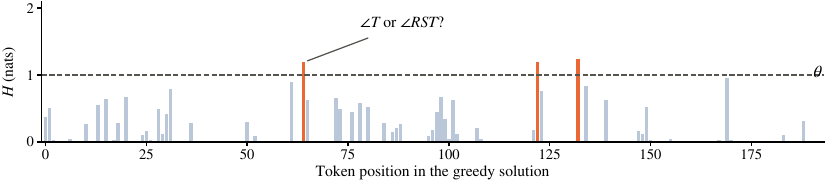}\\[2pt]
\begin{tcolorbox}[enhanced, colback=gray!4, colframe=gray!45, boxrule=0.4pt, arc=1pt, left=4pt, right=4pt, top=3pt, bottom=3pt, fontupper=\footnotesize]
\textbf{Problem} (MATH-500, geometry, level 3). In $\triangle RST$, shown, $\sin R=\frac{2}{5}$. What is $\sin T$? \textit{[Diagram: right angle at $S$; $RT=5$.]}\\[2pt]
\textbf{Common prefix.} \ldots\ Given: \ldots\ $\triangle RST$ is a right triangle with $\angle$\,\colorbox{forkorange!18}{\strut\,?\,}\hfill{\footnotesize\color{gray!70!black}first hesitation point, $H=1.19>\theta$}
\end{tcolorbox}
\begin{tcbraster}[raster columns=2, raster equal height, raster column skip=6pt, raster before skip=2pt]
\begin{tcolorbox}[enhanced, colback=white, colframe=gray!60, boxrule=0.4pt, arc=1pt, left=4pt, right=4pt, top=3pt, bottom=3pt, fontupper=\footnotesize, title={\small Greedy decoding}, coltitle=black, colbacktitle=gray!15]
$\angle$\colorbox{forkorange!18}{$T$}$\,=90^\circ$ \ldots\ ``since $\angle T=90^\circ$, we know that $\sin 90^\circ=1$'' \ldots\ $\boxed{1}$\ \xmark
\end{tcolorbox}
\begin{tcolorbox}[enhanced, colback=white, colframe=hestblue!70, boxrule=0.4pt, arc=1pt, left=4pt, right=4pt, top=3pt, bottom=3pt, fontupper=\footnotesize, title={\small \HEST{}: one geodesic step at this token}, coltitle=black, colbacktitle=hestblue!12]
$\angle$\colorbox{forkorange!18}{$RST$}$\,=90^\circ$ \ldots\ ``$\angle R$ and $\angle T$ are complementary'' \ldots\ $\sin T=\cos R=\boxed{\tfrac{\sqrt{21}}{5}}$\ \cmark
\end{tcolorbox}
\end{tcbraster}
\caption{One steered token changes the course of the solution. Top: next-token entropy along the greedy
solution, in which $3$ of $189$ tokens exceed $\theta=1$. Bottom: the two continuations from
the first of them. \HEST{} steers $12$ of the $583$ tokens of its own solution.
Appendix~\ref{app:case} gives both solutions.}
\label{fig:case}
\end{figure}

\paragraph{An example.}
\label{sec:example}
Figure~\ref{fig:case} follows the MATH-500 problem with the shortest greedy solution among those
that \HEST{} fixes for the development model. The first of its three hesitation points is where
the model states which angle is the right angle. Greedy decoding writes $\angle T$, contrary to
the diagram, and concludes $\sin T=1$. After one geodesic step at this token, the model writes
$\angle RST$ and eventually reaches $\sqrt{21}/5$. In most other fixed problems, the first changed token is
a choice of wording.

\section{Ablation Study}\label{sec:ablation}
We ablate in which direction to steer, in which geometry and where. The first two comparisons
are run in all six settings: every variant keeps the layer, the threshold, the gate, the cache
recomputation and the fallback of \HEST{}, and its edit has the median relative size of the \HEST{}
edit at the selected step length (Table~\ref{tab:ablation}, Appendix~\ref{app:controls}).

\begin{table}[!htb]
\caption{Accuracy (\%) when only the direction of the edit is changed, with the change from
greedy decoding in parentheses. All columns use the gate, the cache recomputation and the fallback
of \HEST{}, and the median relative edit size of \HEST{} at the selected step length. \HEST{} uses the
probe that selects the step length, and \HEST{}, three probes gives the mean and standard deviation over the three probes of Table~\ref{tab:models}. Random: mean and standard deviation over three random
directions. In the last row, the random directions are summed over the three directions. Bold: best per row, without the three-probe mean.}
\label{tab:ablation}
\centering
\small
\setlength{\tabcolsep}{3pt}
\resizebox{\textwidth}{!}{
\begin{tabular}{llccccccc}
\toprule
Model & Dataset & Greedy & \HEST{} & \HEST{}, three probes & Fixed & Euclidean & Reverse & Random \\
\midrule
\multirow{2}{*}{Qwen2.5-Math-1.5B} & MATH-500 & 74.20 & $\mathbf{75.00}$ ($+0.80$) & $74.67\pm0.31$ ($+0.47$) & $74.80$ ($+0.60$) & $73.80$ ($-0.40$) & $74.20$ ($0.00$) & $74.20\pm0.35$ ($0.00$) \\
 & GSM8K & 84.84 & $\mathbf{85.29}$ ($+0.45$) & $85.01\pm0.31$ ($+0.17$) & $84.76$ ($-0.08$) & $84.69$ ($-0.15$) & $84.46$ ($-0.38$) & $84.56\pm0.63$ ($-0.28$) \\
\midrule
\multirow{2}{*}{Llama-3.1-8B} & MATH-500 & 50.80 & $\mathbf{52.00}$ ($+1.20$) & $51.87\pm0.12$ ($+1.07$) & $49.40$ ($-1.40$) & $50.40$ ($-0.40$) & $50.20$ ($-0.60$) & $50.20\pm0.92$ ($-0.60$) \\
 & GSM8K & 84.61 & $\mathbf{85.06}$ ($+0.45$) & $84.69\pm0.42$ ($+0.08$) & $84.23$ ($-0.38$) & $84.08$ ($-0.53$) & $84.31$ ($-0.30$) & $84.53\pm0.23$ ($-0.08$) \\
\midrule
\multirow{2}{*}{Qwen2.5-Math-7B} & MATH-500 & 84.00 & $\mathbf{84.60}$ ($+0.60$) & $85.07\pm0.42$ ($+1.07$) & $83.80$ ($-0.20$) & $84.20$ ($+0.20$) & $82.20$ ($-1.80$) & $84.33\pm0.12$ ($+0.33$) \\
 & GSM8K & 95.38 & $\mathbf{95.68}$ ($+0.30$) & $95.55\pm0.16$ ($+0.17$) & $95.53$ ($+0.15$) & $95.38$ ($0.00$) & $95.30$ ($-0.08$) & $95.58\pm0.36$ ($+0.20$) \\
\midrule
\multicolumn{2}{l}{Mean over the six settings} & 78.97 & $\mathbf{79.61}$ ($+0.64$) & $79.48$ ($+0.50$) & $78.75$ ($-0.22$) & $78.76$ ($-0.21$) & $78.44$ ($-0.53$) & $78.90$ ($-0.07$) \\
\multicolumn{3}{l}{Number of problems fixed / broken} & $93/64$ & -- & $80/89$ & $73/85$ & $69/91$ & $224/234$ \\
\bottomrule
\end{tabular}}
\end{table}

\paragraph{Which direction.}
\emph{Fixed} uses the unit CAA direction of Section~\ref{sec:setup}, \emph{reverse} steps away from
the ideal point with $-t$ in place of $t$, and \emph{random} uses a random unit vector of the
hidden space, drawn once per run. Averaged over the six settings, only \HEST{} improves on greedy
decoding, by $0.64$ points, and it is the only variant that fixes more problems than it breaks.
The fixed direction changes the average by $-0.22$ points, the random directions by $-0.07$ and
the reverse direction by $-0.53$. \HEST{} has the highest accuracy in all six settings. With the mean over the three probes, \HEST{} gains $0.50$ points on average and stays above greedy decoding in all six settings, whereas the fixed direction is below it in four, the random directions in three and the reverse direction in five. The difference is clearest for
Llama-3.1-8B-Instruct on MATH-500, where \HEST{} gains $1.20$ points and the fixed direction loses
$1.40$. A small edit at hesitation points is thus close to neutral in most directions, and the
direction of the probe is the one that gives a consistent gain.

\paragraph{Which geometry.}
The Euclidean probe, the limit $c\to0$ of the probe of \HEST{}, omits the exponential map and is
trained in the same way. Its step
length matches the median edit size of \HEST{} on training states (Appendix~\ref{app:euclid}).
With the Busemann readout, it changes accuracy by $-0.53$ to $+0.20$ points and by $-0.21$ points
on average, and it is below \HEST{} in all six settings. Over the six settings it
fixes $73$ problems and breaks $85$. The curvature thus matters for the Busemann readout, as
Section~\ref{sec:step} predicts.

\paragraph{Where to steer.}
For the development model on MATH-500, at $t=0.5$, we also apply a step of the same size along
the probe direction at other positions (Table~\ref{tab:positions}). \emph{Every token} edits all generated positions, and \emph{step boundaries} edits the tokens that
contain a double line break, where the reasoning steps of SEAL begin. Both variants fall below
greedy decoding, by $0.20$ and $0.40$ points, whereas the same step at hesitation points gains
$1.80$ points at this step length. Steering step boundaries
edits about as many tokens as \HEST{}, but only $8.2\%$ of the hesitation points fall at the
start of a step. CAA adds a vector about five times larger than the \HEST{} step at every
token and lowers accuracy by $1.80$ points. Appendix~\ref{app:ablation} also varies the threshold
and the layer, and neither improves on $\theta=1$ and layer $19$.

\paragraph{Step length.}
For the development model, we extend the sweep on the MATH validation set to $t=2$ and $t=5$
(Appendix~\ref{app:mechanism}). The number of solved problems is highest at the selected $t=0.25$
and decreases slowly, from $416$ to $402$ of $500$ at $t=5$. At this step length, the edit is almost
half as large as the hidden state and changes the most likely token at $39\%$ of the hesitation
points, against $5\%$ at $t=0.25$. \HEST{} thus remains stable over a wide range of step lengths,
because it edits few positions.

\begin{table}[!htb]
\caption{Where to steer, on MATH-500 with Qwen2.5-Math-1.5B-Instruct and the Busemann readout at
$t=0.5$, with one probe. Accuracy in \% and change from greedy decoding in points. Steered: median
fraction of steered tokens. Step: median $\lVert\Delta h\rVert/\lVert h\rVert$.}
\label{tab:positions}
\centering
\small
\begin{tabular}{lccc}
\toprule
Positions & Steered & Step & Accuracy \\
\midrule
Hesitation points (\HEST{}) & $1.1\%$ & $0.06$ & $\mathbf{76.00}$ ($+1.80$) \\
Every token & $100\%$ & $0.07$ & $74.00$ ($-0.20$) \\
Step boundaries & $1.4\%$ & $0.07$ & $73.80$ ($-0.40$) \\
\midrule
CAA, every token & $100\%$ & $0.33$ & $72.40$ ($-1.80$) \\
\bottomrule
\end{tabular}
\end{table}

\section{Limitations and Conclusion}\label{sec:conclusion}
\paragraph{Limitations.}
\HEST{} acts on the problems that can still be fixed, so its gains are largest where greedy
decoding leaves room and on solutions with distinct hesitation points. Benchmarks on which the
model hesitates densely, such as OlympiadBench (Appendix~\ref{app:fixedt}), other decoding methods, models with more than $8$B
parameters and tasks beyond mathematical reasoning are natural next steps.

\paragraph{Conclusion.}
\HEST{} steers a language model with hyperbolic geometry, using the model's own next-token
entropy as the only label of its probe. The entropy
selects the few tokens at which the model hesitates, and a probe trained on the same signal gives
the direction through the Riemannian gradient of its readout. Steering only these tokens improves
greedy accuracy on mathematical reasoning on average in every setting and on the full MATH test
set, whereas a fixed vector added at every
token lowers it. With the same gate and edit size, a Euclidean probe that is trained in the
same way, a fixed direction and random directions do not give this gain, which shows that it
comes from the direction of the hyperbolic probe.
The code is available at \url{https://github.com/Zeyong98/HEST}.

\section*{Acknowledgments}

This work was supported by the DOE SEA-CROGS project (DE-SC0023191) and the AFOSR project
(FA9550-24-1-0231).

\bibliography{references}

\begin{thebibliography}{51}
\providecommand{\natexlab}[1]{#1}
\providecommand{\url}[1]{\texttt{#1}}
\expandafter\ifx\csname urlstyle\endcsname\relax
  \providecommand{\doi}[1]{doi: #1}\else
  \providecommand{\doi}{doi: \begingroup \urlstyle{rm}\Url}\fi

\bibitem[Atkinson(2026)]{atkinson2026tethered}
Craig Atkinson.
\newblock Tethered reasoning: Decoupling entropy from hallucination in
  quantized {LLMs} via manifold steering.
\newblock \emph{arXiv preprint arXiv:2602.17691}, 2026.

\bibitem[Bridson and Haefliger(2013)]{bridson2013metric}
Martin~R Bridson and Andr{\'e} Haefliger.
\newblock \emph{Metric spaces of non-positive curvature}.
\newblock Springer Science \& Business Media, 2013.

\bibitem[Chen et~al.(2025)Chen, Zhang, Hong, Kundu, and Wang]{chen2025seal}
Runjin Chen, Zhenyu Zhang, Junyuan Hong, Souvik Kundu, and Zhangyang Wang.
\newblock {SEAL}: Steerable reasoning calibration of large language models for
  free.
\newblock In \emph{Second Conference on Language Modeling}, 2025.
\newblock URL \url{https://openreview.net/forum?id=klPszYDIRT}.

\bibitem[Chen et~al.(2026)Chen, Sch{\"o}lkopf, and Sebe]{chen2026hbnn}
Ziheng Chen, Bernhard Sch{\"o}lkopf, and Nicu Sebe.
\newblock Hyperbolic {Busemann} neural networks.
\newblock In \emph{CVPR}, 2026.

\bibitem[Cheng et~al.(2026)Cheng, Gan, Jiang, Wang, Yin, Luo, Fu, and
  Gu]{cheng2026steering}
Zifeng Cheng, Jinwei Gan, Zhiwei Jiang, Cong Wang, Yafeng Yin, Xiang Luo,
  Yuchen Fu, and Qing Gu.
\newblock Steering when necessary: Flexible steering large language models with
  backtracking.
\newblock \emph{Advances in Neural Information Processing Systems},
  38:\penalty0 8390--8413, 2026.

\bibitem[Cobbe et~al.(2021)Cobbe, Kosaraju, Bavarian, Chen, Jun, Kaiser,
  Plappert, Tworek, Hilton, Nakano, et~al.]{cobbe2021training}
Karl Cobbe, Vineet Kosaraju, Mohammad Bavarian, Mark Chen, Heewoo Jun, Lukasz
  Kaiser, Matthias Plappert, Jerry Tworek, Jacob Hilton, Reiichiro Nakano,
  et~al.
\newblock Training verifiers to solve math word problems.
\newblock \emph{arXiv preprint arXiv:2110.14168}, 2021.

\bibitem[Ganea et~al.(2018)Ganea, B{\'e}cigneul, and
  Hofmann]{ganea2018hyperbolic}
Octavian Ganea, Gary B{\'e}cigneul, and Thomas Hofmann.
\newblock Hyperbolic neural networks.
\newblock \emph{Advances in Neural Information Processing Systems}, 31, 2018.

\bibitem[Ghadimi~Atigh et~al.(2021)Ghadimi~Atigh, Keller-Ressel, and
  Mettes]{ghadimi2021hyperbolic}
Mina Ghadimi~Atigh, Martin Keller-Ressel, and Pascal Mettes.
\newblock Hyperbolic {Busemann} learning with ideal prototypes.
\newblock \emph{Advances in Neural Information Processing Systems},
  34:\penalty0 103--115, 2021.

\bibitem[Grattafiori et~al.(2024)Grattafiori, Dubey, Jauhri, Pandey, Kadian,
  Al-Dahle, Letman, Mathur, Schelten, Vaughan, et~al.]{grattafiori2024llama}
Aaron Grattafiori, Abhimanyu Dubey, Abhinav Jauhri, Abhinav Pandey, Abhishek
  Kadian, Ahmad Al-Dahle, Aiesha Letman, Akhil Mathur, Alan Schelten, Alex
  Vaughan, et~al.
\newblock The {Llama} 3 herd of models.
\newblock \emph{arXiv preprint arXiv:2407.21783}, 2024.

\bibitem[He et~al.(2024)He, Luo, Bai, Hu, Thai, Shen, Hu, Han, Huang, Zhang,
  et~al.]{he2024olympiadbench}
Chaoqun He, Renjie Luo, Yuzhuo Bai, Shengding Hu, Zhen Thai, Junhao Shen, Jinyi
  Hu, Xu~Han, Yujie Huang, Yuxiang Zhang, et~al.
\newblock {OlympiadBench}: A challenging benchmark for promoting {AGI} with
  olympiad-level bilingual multimodal scientific problems.
\newblock In \emph{Proceedings of the 62nd Annual Meeting of the Association
  for Computational Linguistics (Volume 1: Long Papers)}, pages 3828--3850,
  2024.

\bibitem[He et~al.(2026)He, Anand, Madhu, Maatouk, Krishnaswamy, Tassiulas,
  Yang, and Ying]{he2026helm}
Neil He, Rishabh Anand, Hiren Madhu, Ali Maatouk, Smita Krishnaswamy, Leandros
  Tassiulas, Menglin Yang, and Rex Ying.
\newblock {HELM}: Hyperbolic large language models via mixture-of-curvature
  experts.
\newblock \emph{Advances in Neural Information Processing Systems},
  38:\penalty0 142604--142635, 2026.

\bibitem[Hendrycks et~al.(2021)Hendrycks, Burns, Kadavath, Arora, Basart, Tang,
  Song, and Steinhardt]{hendrycks2021measuring}
Dan Hendrycks, Collin Burns, Saurav Kadavath, Akul Arora, Steven Basart, Eric
  Tang, Dawn Song, and Jacob Steinhardt.
\newblock Measuring mathematical problem solving with the {MATH} dataset.
\newblock \emph{arXiv preprint arXiv:2103.03874}, 2021.

\bibitem[Herbrich et~al.(2000)Herbrich, Graepel, and
  Obermayer]{herbrich2000large}
Ralf Herbrich, Thore Graepel, and Klaus Obermayer.
\newblock Large margin rank boundaries for ordinal regression.
\newblock In Alexander~J. Smola, Peter Bartlett, Bernhard Sch{\"o}lkopf, and
  Dale Schuurmans, editors, \emph{Advances in Large Margin Classifiers}, pages
  115--132. MIT Press, 2000.
\newblock \doi{10.7551/mitpress/1113.003.0010}.

\bibitem[Hsu et~al.(2026)Hsu, Beaglehole, Radhakrishnan, and
  Belkin]{hsu2026contextual}
Brandon Hsu, Daniel Beaglehole, Adityanarayanan Radhakrishnan, and Mikhail
  Belkin.
\newblock Contextual linear activation steering of language models.
\newblock \emph{arXiv preprint arXiv:2604.24693}, 2026.

\bibitem[Jin et~al.(2026)Jin, Deng, Wang, Shen, and Zhang]{jin2026beyond}
Zehao Jin, Ruixuan Deng, Junran Wang, Xinjie Shen, and Chao Zhang.
\newblock Beyond steering vector: Flow-based activation steering for
  inference-time intervention.
\newblock In \emph{Mechanistic Interpretability Workshop at ICML 2026}, 2026.
\newblock URL \url{https://openreview.net/forum?id=t7M7vjIJbe}.

\bibitem[Khanh et~al.(2026)Khanh, Zhu, Yue, and Nguyen]{khanh2026test}
Ly~Tran~Ho Khanh, Dongxuan Zhu, Man-Chung Yue, and Viet~Anh Nguyen.
\newblock Test-time diverse reasoning by {Riemannian} activation steering.
\newblock In \emph{Proceedings of the AAAI Conference on Artificial
  Intelligence}, volume~40, pages 31429--31437, 2026.

\bibitem[Kim et~al.(2026)Kim, Lee, Yeom, Sok, Park, Park, Kim, and
  Kim]{kim2026noise}
Heejun Kim, Seungpil Lee, Jewon Yeom, Jaewon Sok, Seonghyeon Park, Jeongjae
  Park, Taesup Kim, and Sundong Kim.
\newblock From noise to diversity: Random embedding injection in {LLM}
  reasoning.
\newblock \emph{arXiv preprint arXiv:2605.11936}, 2026.

\bibitem[Kydl{\'i}{\v{c}}ek(2025)]{kydlicek2025mathverify}
Hynek Kydl{\'i}{\v{c}}ek.
\newblock {Math-Verify}: Math verification library, 2025.
\newblock URL \url{https://github.com/huggingface/math-verify}.

\bibitem[Li et~al.(2025)Li, Li, Wu, Zhu, Wang, Yu, Jiang, Zhu, Jia, Wu,
  et~al.]{li2025seek}
Hengli Li, Chenxi Li, Tong Wu, Xuekai Zhu, Yuxuan Wang, Zhaoxin Yu,
  Eric~Hanchen Jiang, Song-Chun Zhu, Zixia Jia, Ying~Nian Wu, et~al.
\newblock Seek in the dark: Reasoning via test-time instance-level policy
  gradient in latent space.
\newblock \emph{arXiv preprint arXiv:2505.13308}, 2025.

\bibitem[Li et~al.(2023)Li, Patel, Vi{\'e}gas, Pfister, and
  Wattenberg]{li2023inference}
Kenneth Li, Oam Patel, Fernanda Vi{\'e}gas, Hanspeter Pfister, and Martin
  Wattenberg.
\newblock Inference-time intervention: Eliciting truthful answers from a
  language model.
\newblock \emph{Advances in Neural Information Processing Systems},
  36:\penalty0 41451--41530, 2023.

\bibitem[Li et~al.(2026{\natexlab{a}})Li, Callanan, Ghassel, and
  Zhu]{li2026entropy}
Xianzhi Li, Ethan Callanan, Abdellah Ghassel, and Xiaodan Zhu.
\newblock Entropy-gated branching for efficient test-time reasoning.
\newblock In \emph{Proceedings of the 19th Conference of the European Chapter
  of the Association for Computational Linguistics (Volume 1: Long Papers)},
  pages 5054--5069, 2026{\natexlab{a}}.

\bibitem[Li et~al.(2026{\natexlab{b}})Li, Bergner, Zhao, Patil, Chen, and
  Wang]{li2026steering}
Yawei Li, Benjamin Bergner, Yinghan Zhao, Vihang~Prakash Patil, Bei Chen, and
  Cheng Wang.
\newblock Steering large reasoning models towards concise reasoning via flow
  matching.
\newblock \emph{Transactions on Machine Learning Research}, 2026{\natexlab{b}}.
\newblock URL \url{https://openreview.net/forum?id=qwcJMdGerK}.

\bibitem[Lightman et~al.(2024)Lightman, Kosaraju, Burda, Edwards, Baker, Lee,
  Leike, Schulman, Sutskever, and Cobbe]{lightman2024let}
Hunter Lightman, Vineet Kosaraju, Yuri Burda, Harrison Edwards, Bowen Baker,
  Teddy Lee, Jan Leike, John Schulman, Ilya Sutskever, and Karl Cobbe.
\newblock Let's verify step by step.
\newblock In \emph{International Conference on Learning Representations},
  volume 2024, pages 39578--39601, 2024.

\bibitem[Liu et~al.(2025)Liu, Chen, Lu, Ye, Chen, Xing, and
  Zou]{liu2025fractional}
Sheng Liu, Tianlang Chen, Pan Lu, Haotian Ye, Yizheng Chen, Lei Xing, and James
  Zou.
\newblock Fractional reasoning via latent steering vectors improves inference
  time compute.
\newblock \emph{arXiv preprint arXiv:2506.15882}, 2025.

\bibitem[Liu et~al.(2026)Liu, Xu, He, Patil, Xu, and Ma]{liu2026hyperguide}
Yuyu Liu, Haotian Xu, Yanan He, Sarang~Rajendra Patil, Mengjia Xu, and Tengfei
  Ma.
\newblock {HyperGuide}: Hyperbolic guidance for efficient multi-step reasoning
  in large language models.
\newblock \emph{arXiv preprint arXiv:2605.24140}, 2026.

\bibitem[Nguyen et~al.(2025)Nguyen, Prasad, Stengel-Eskin, and
  Bansal]{nguyen2025multi}
Duy Nguyen, Archiki Prasad, Elias Stengel-Eskin, and Mohit Bansal.
\newblock Multi-attribute steering of language models via targeted
  intervention.
\newblock In \emph{Proceedings of the 63rd Annual Meeting of the Association
  for Computational Linguistics (Volume 1: Long Papers)}, pages 20619--20634,
  2025.

\bibitem[Nickel and Kiela(2017)]{nickel2017poincare}
Maximillian Nickel and Douwe Kiela.
\newblock Poincar{\'e} embeddings for learning hierarchical representations.
\newblock \emph{Advances in Neural Information Processing Systems}, 30, 2017.

\bibitem[Oozeer et~al.(2025)Oozeer, Marks, Barez, and
  Abdullah]{oozeer-etal-2025-beyond}
Narmeen~Fatimah Oozeer, Luke Marks, Fazl Barez, and Amir Abdullah.
\newblock Beyond linear steering: Unified multi-attribute control for language
  models.
\newblock In Christos Christodoulopoulos, Tanmoy Chakraborty, Carolyn Rose, and
  Violet Peng, editors, \emph{Findings of the Association for Computational
  Linguistics: EMNLP 2025}, pages 23513--23557, Suzhou, China, 2025.
  Association for Computational Linguistics.
\newblock \doi{10.18653/v1/2025.findings-emnlp.1278}.
\newblock URL \url{https://aclanthology.org/2025.findings-emnlp.1278/}.

\bibitem[Park et~al.(2025)Park, Choe, Jiang, and Veitch]{park2025geometry}
Kiho Park, Yo~Joong Choe, Yibo Jiang, and Victor Veitch.
\newblock The geometry of categorical and hierarchical concepts in large
  language models.
\newblock In \emph{International Conference on Learning Representations},
  volume 2025, pages 76441--76463, 2025.

\bibitem[Patil et~al.(2026{\natexlab{a}})Patil, Zhang, Huang, Ma, and
  Xu]{patil2026hyperbolic}
Sarang Patil, Zeyong Zhang, Yiran Huang, Tengfei Ma, and Mengjia Xu.
\newblock Hyperbolic large language models.
\newblock \emph{SIAM Review}, 68\penalty0 (4):\penalty0 885--913,
  2026{\natexlab{a}}.

\bibitem[Patil et~al.(2026{\natexlab{b}})Patil, Pandey, Koutis, and
  Xu]{patil2026hierarchical}
Sarang~Rajendra Patil, Ashish~Parmanand Pandey, Ioannis Koutis, and Mengjia Xu.
\newblock Hierarchical {Mamba} meets hyperbolic geometry: A new paradigm for
  structured language embeddings.
\newblock \emph{Transactions on Machine Learning Research}, 2026{\natexlab{b}}.
\newblock URL \url{https://openreview.net/forum?id=a3g13FKzct}.

\bibitem[Raval et~al.(2026)Raval, Song, Wu, Harrasse, Phillips, Barez, and
  Abdullah]{raval2026curveball}
Shivam Raval, Hae~Jin Song, Linlin Wu, Abir Harrasse, Jeff~M Phillips, Fazl
  Barez, and Amirali Abdullah.
\newblock Curveball steering: The right direction to steer isn't always linear.
\newblock \emph{arXiv preprint arXiv:2603.09313}, 2026.

\bibitem[Rimsky et~al.(2024)Rimsky, Gabrieli, Schulz, Tong, Hubinger, and
  Turner]{rimsky2024steering}
Nina Rimsky, Nick Gabrieli, Julian Schulz, Meg Tong, Evan Hubinger, and
  Alexander Turner.
\newblock Steering {Llama} 2 via contrastive activation addition.
\newblock In \emph{Proceedings of the 62nd Annual Meeting of the Association
  for Computational Linguistics (Volume 1: Long Papers)}, pages 15504--15522,
  2024.

\bibitem[Rodriguez et~al.(2025)Rodriguez, Blaas, Klein, Zappella, Apostoloff,
  Suau, et~al.]{rodriguez2025controlling}
Pau Rodriguez, Arno Blaas, Michal Klein, Luca Zappella, Nicholas Apostoloff,
  Xavier Suau, et~al.
\newblock Controlling language and diffusion models by transporting
  activations.
\newblock In \emph{International Conference on Learning Representations},
  volume 2025, pages 89812--89855, 2025.

\bibitem[Sala et~al.(2018)Sala, De~Sa, Gu, and R{\'e}]{sala2018representation}
Frederic Sala, Chris De~Sa, Albert Gu, and Christopher R{\'e}.
\newblock Representation tradeoffs for hyperbolic embeddings.
\newblock In \emph{International Conference on Machine Learning}, pages
  4460--4469. PMLR, 2018.

\bibitem[Schroff et~al.(2015)Schroff, Kalenichenko, and
  Philbin]{schroff2015facenet}
Florian Schroff, Dmitry Kalenichenko, and James Philbin.
\newblock {FaceNet}: A unified embedding for face recognition and clustering.
\newblock In \emph{Proceedings of the IEEE Conference on Computer Vision and
  Pattern Recognition}, pages 815--823, 2015.

\bibitem[Stora{\"\i} and Hwang(2025)]{storai2025harp}
Romain Stora{\"\i} and Seung-won Hwang.
\newblock {HARP}: Hesitation-aware reframing in transformer inference pass.
\newblock In \emph{Proceedings of the 2025 Conference of the Nations of the
  Americas Chapter of the Association for Computational Linguistics: Human
  Language Technologies (Volume 1: Long Papers)}, pages 12305--12319, 2025.

\bibitem[Sun et~al.(2025)Sun, Baskaran, Wu, Sklar, Potts, and
  Geiger]{sun2025hypersteer}
Jiuding Sun, Sidharth Baskaran, Zhengxuan Wu, Michael Sklar, Christopher Potts,
  and Atticus Geiger.
\newblock {HyperSteer}: Activation steering at scale with hypernetworks.
\newblock \emph{arXiv preprint arXiv:2506.03292}, 2025.

\bibitem[Sun et~al.(2026)Sun, Dong, Qiao, Lin, Zhang, and Rajmohan]{sun2026llm}
Lihao Sun, Hang Dong, Bo~Qiao, Qingwei Lin, Dongmei Zhang, and Saravan
  Rajmohan.
\newblock {LLM} reasoning as trajectories: Step-specific representation
  geometry and correctness signals.
\newblock In \emph{Proceedings of the 64th Annual Meeting of the Association
  for Computational Linguistics (Volume 1: Long Papers)}, pages 26872--26887,
  2026.

\bibitem[Turner et~al.(2023)Turner, Thiergart, Leech, Udell, Vazquez, Mini, and
  MacDiarmid]{turner2023steering}
Alexander~Matt Turner, Lisa Thiergart, Gavin Leech, David Udell, Juan~J
  Vazquez, Ulisse Mini, and Monte MacDiarmid.
\newblock Steering language models with activation engineering.
\newblock \emph{arXiv preprint arXiv:2308.10248}, 2023.

\bibitem[Vu and Nguyen(2026)]{vu2026angular}
Minh~Hieu Vu and Tan Nguyen.
\newblock Angular steering: Behavior control via rotation in activation space.
\newblock \emph{Advances in Neural Information Processing Systems},
  38:\penalty0 121653--121690, 2026.

\bibitem[Wang et~al.(2026)Wang, Yu, Gao, Zheng, Liu, Lu, Dang, Chen, Yang,
  Zhang, et~al.]{wang2026beyond}
Shenzhi Wang, Le~Yu, Chang Gao, Chujie Zheng, Shixuan Liu, Rui Lu, Kai Dang,
  Xiong-Hui Chen, Jianxin Yang, Zhenru Zhang, et~al.
\newblock Beyond the 80/20 rule: High-entropy minority tokens drive effective
  reinforcement learning for {LLM} reasoning.
\newblock \emph{Advances in Neural Information Processing Systems},
  38:\penalty0 115452--115486, 2026.

\bibitem[Weinberger and Saul(2009)]{weinberger2009distance}
Kilian~Q Weinberger and Lawrence~K Saul.
\newblock Distance metric learning for large margin nearest neighbor
  classification.
\newblock \emph{Journal of Machine Learning Research}, 10\penalty0 (2), 2009.

\bibitem[Wu et~al.(2024)Wu, Arora, Wang, Geiger, Jurafsky, Manning, and
  Potts]{wu2024reft}
Zhengxuan Wu, Aryaman Arora, Zheng Wang, Atticus Geiger, Dan Jurafsky,
  Christopher~D Manning, and Christopher Potts.
\newblock {ReFT}: Representation finetuning for language models.
\newblock \emph{Advances in Neural Information Processing Systems},
  37:\penalty0 63908--63962, 2024.

\bibitem[Yang et~al.(2024{\natexlab{a}})Yang, Zhang, Hui, Gao, Yu, Li, Liu, Tu,
  Zhou, Lin, et~al.]{yang2024qwen2}
An~Yang, Beichen Zhang, Binyuan Hui, Bofei Gao, Bowen Yu, Chengpeng Li,
  Dayiheng Liu, Jianhong Tu, Jingren Zhou, Junyang Lin, et~al.
\newblock {Qwen2.5-Math} technical report: Toward mathematical expert model via
  self-improvement.
\newblock \emph{arXiv preprint arXiv:2409.12122}, 2024{\natexlab{a}}.

\bibitem[Yang et~al.(2024{\natexlab{b}})Yang, Verma, Zhang, Liu, King, and
  Ying]{yang2024hypformer}
Menglin Yang, Harshit Verma, Delvin~Ce Zhang, Jiahong Liu, Irwin King, and Rex
  Ying.
\newblock Hypformer: Exploring efficient transformer fully in hyperbolic space.
\newblock In \emph{Proceedings of the 30th ACM SIGKDD Conference on Knowledge
  Discovery and Data Mining}, pages 3770--3781, 2024{\natexlab{b}}.

\bibitem[Yang et~al.(2026)Yang, Feng, Xiong, Liu, King, Ying,
  et~al.]{yang2026hyperbolic}
Menglin Yang, Aosong Feng, Bo~Xiong, Jiahong Liu, Irwin King, Rex Ying, et~al.
\newblock Hyperbolic fine-tuning for large language models.
\newblock \emph{Advances in Neural Information Processing Systems},
  38:\penalty0 29613--29648, 2026.

\bibitem[You et~al.(2026)You, Deng, and Chen]{you2026spherical}
Zejia You, Chunyuan Deng, and Hanjie Chen.
\newblock Spherical steering: Geometry-aware activation rotation for language
  models.
\newblock In \emph{Forty-third International Conference on Machine Learning},
  2026.
\newblock URL \url{https://openreview.net/forum?id=9PPeJsVLdB}.

\bibitem[Zhao et~al.(2026)Zhao, Sun, Kong, Li, Wang, Jiang, Zhu, Abdelzaher,
  Choi, Li, and Shao]{zhao2026odesteer}
Hongjue Zhao, Haosen Sun, Jiangtao Kong, Xiaochang Li, Qineng Wang, Liwei
  Jiang, Qi~Zhu, Tarek~F. Abdelzaher, Yejin Choi, Manling Li, and Huajie Shao.
\newblock {ODES}teer: A unified {ODE}-based steering framework for {LLM}
  alignment.
\newblock In \emph{The Fourteenth International Conference on Learning
  Representations}, 2026.
\newblock URL \url{https://openreview.net/forum?id=CFewUmgIIL}.

\bibitem[Zou et~al.(2023)Zou, Phan, Chen, Campbell, Guo, Ren, Pan, Yin,
  Mazeika, Dombrowski, et~al.]{zou2023representation}
Andy Zou, Long Phan, Sarah Chen, James Campbell, Phillip Guo, Richard Ren,
  Alexander Pan, Xuwang Yin, Mantas Mazeika, Ann-Kathrin Dombrowski, et~al.
\newblock Representation engineering: A top-down approach to {AI} transparency.
\newblock \emph{arXiv preprint arXiv:2310.01405}, 2023.

\bibitem[Zur et~al.(2025)Zur, Geiger, Lubana, and Bigelow]{zur2025language}
Amir Zur, Atticus Geiger, Ekdeep~Singh Lubana, and Eric Bigelow.
\newblock Are language models aware of the road not taken? {T}oken-level
  uncertainty and hidden state dynamics.
\newblock \emph{arXiv preprint arXiv:2511.04527}, 2025.

\end{thebibliography}
\bibliographystyle{plainnat}

\clearpage
\appendix

\begin{center}
{\Large\bfseries Appendix}
\end{center}
\vspace{2pt}
\startcontents[appendix]
\printcontents[appendix]{l}{1}{\setcounter{tocdepth}{1}}
\vspace{6pt}

\section{Comparison with Related Steering Methods}
\label{app:position}

Table~\ref{tab:app-position} compares \HEST{} with the steering methods discussed in
Section~\ref{sec:related-work} along the design choices studied in this paper. Most methods edit
every token or fixed positions of the prompt, and MAT-Steer weights its edits of the prompt tokens
with a learned gate. The few methods that restrict the positions during generation use a
confidence score computed from a contrastive direction, a probe trained on labels, step
boundaries, or a score that combines the next-token entropy with a distance to a reference
distribution. Helix uses the last of these scores to lower the largest
logit and does not edit the hidden state. \HEST{} restricts the positions with the entropy alone,
and the same entropy trains the probe that gives the direction. Most methods also derive their
direction from contrastive or labeled data. Among the methods that report accuracy on
mathematical benchmarks, SEAL and FlowSteer target long reasoning models, and the method of
\citet{sun2026llm} needs correctness labels. CLAS uses GSM8K to elicit step-by-step solutions,
ReFT trains its interventions on the task, Helix evaluates a quantized model at high sampling
temperatures, and Fractional Reasoning and SPREAD report accuracy over several samples.

\begin{table}[!t]
\caption{Steering methods compared along the design choices of \HEST{}. \emph{Positions}: the
tokens at which the edit is applied. \emph{Math}: whether accuracy on mathematical benchmarks is
reported, and in which setting. Each entry was checked against the original paper.}
\label{tab:app-position}
\centering
\scriptsize
\setlength{\tabcolsep}{3pt}
\renewcommand{\arraystretch}{1.12}
\begin{tabular}{p{1.9cm}p{1.9cm}p{1.9cm}p{1.8cm}p{1.1cm}p{2.2cm}p{1.5cm}}
\toprule
Method & Direction & Positions & Training signal & Geometry & Evaluation & Math \\
\midrule
CAA \citep{rimsky2024steering} & fixed, difference of means & all tokens after the prompt & contrastive pairs & Euclidean & behavioral questions, open-ended generation, MMLU, TruthfulQA & -- \\
ITI \citep{li2023inference} & fixed, top attention heads & every generated token & labeled truthful and untruthful answers & Euclidean & TruthfulQA, NQ, TriviaQA, MMLU & -- \\
CLAS \citep{hsu2026contextual} & per-layer direction, coefficient linear in the activation & all tokens & probe direction, prompt and completion pairs & Euclidean & $11$ tasks including GSM8K & GSM8K, to elicit step-by-step solutions \\
\midrule
ReFT \citep{wu2024reft} & learned low-rank edit of a subspace & fixed prefix and suffix positions of the prompt & task supervision, frozen model & Euclidean & commonsense and arithmetic reasoning, instruction following, GLUE & GSM8K and three others, after task finetuning \\
HyperSteer \citep{sun2025hypersteer} & generated by a hypernetwork from the steering prompt and the input & all tokens & AxBench steering prompts, language-modeling loss & Euclidean & AxBench concepts & -- \\
AcT \citep{rodriguez2025controlling} & coordinate-wise affine maps from optimal transport & all tokens & source and target sentence sets & Euclidean & toxicity, concept induction, TruthfulQA, text-to-image models & -- \\
ODESteer \citep{zhao2026odesteer} & classifier gradient, integrated as an ODE & all generated tokens & positive and negative examples & Euclidean & UltraFeedback, TruthfulQA, RealToxicityPrompts & -- \\
FLAS \citep{jin2026beyond} & concept-conditioned velocity field & all tokens & positive examples, language-modeling loss & Euclidean & AxBench concepts & -- \\
Curveball \citep{raval2026curveball} & fixed direction in kernel PCA space & all generated tokens & contrastive pairs & kernel feature space & behavioral choices, tone traits & -- \\
K-Steering \citep{oozeer-etal-2025-beyond} & classifier gradient & all tokens & attribute labels & Euclidean & tone and debate style, MMLU as a capability check & -- \\
Angular \citep{vu2026angular} & rotation in a fixed plane & all tokens; adaptive variant where aligned with the feature & contrastive prompt sets & sphere & refusal, emotion & tinyGSM8K, as a capability check \\
Spherical \citep{you2026spherical} & geodesic rotation towards a fixed direction & each token, strength set by a confidence gate & contrastive answer pairs & sphere & TruthfulQA and five multiple-choice benchmarks & -- \\
\midrule
MAT-Steer \citep{nguyen2025multi} & fixed per attribute & learned soft gate per prompt token & labeled positive and negative samples & Euclidean, norm preserving & TruthfulQA, Toxigen, BBQ, HelpSteer & -- \\
FASB \citep{cheng2026steering} & fixed per head & probe-triggered, with backtracking & labeled probe & Euclidean & TruthfulQA, six multiple-choice tasks & -- \\
Helix \citep{atkinson2026tethered} & penalty on the largest logit, no edit of the hidden state & tokens with low confidence from entropy and Mahalanobis distance & activations of low-temperature generations on factual prompts & Euclidean, Mahalanobis & GSM8K, HumanEval, MMLU, creative ideation & GSM8K, 4-bit model, temperatures $0.5$ to $3.0$ \\
SEAL \citep{chen2025seal} & fixed & step boundaries & steps classified by keywords & Euclidean & MATH-500, GSM8K, LiveCodeBench & long reasoning models \\
FlowSteer \citep{li2026steering} & flow matching & step boundaries & as SEAL & Euclidean & MATH-500, GSM8K, AIME, AMC, LiveCodeBench & long reasoning models \\
\citet{sun2026llm} & fixed, or towards a centroid & step boundaries or before the answer, gated & correctness labels & Euclidean & GSM8K, MATH-500, MMLU & Llama-3.1-8B-Instruct \\
\midrule
Fractional Reasoning \citep{liu2025fractional} & fixed, scaled, norm preserved & all tokens & contrastive instruction prompts & Euclidean & GSM8K, MATH-500, GPQA & with voting or best-of-$N$ \\
SPREAD \citep{khanh2026test} & optimized jointly for $N$ samples & tokens after each synchronization anchor & none & Euclidean, fixed vector norms & AIME, MATH-500, OlympiadBench & Pass@$N$ \\
\midrule
\textbf{\HEST{}} (ours) & Riemannian gradient of a probe readout & hesitation points & the model's own entropy & hyperbolic & MATH-500, GSM8K, OlympiadBench & three instruction-tuned models \\
\bottomrule
\end{tabular}
\end{table}

\paragraph{State-dependent methods as baselines.}
We do not run the state-dependent methods of Table~\ref{tab:app-position} as baselines.
K-Steering, ODESteer, FLAS and CLAS learn their edits from attribute labels, contrastive examples or
positive examples of a target behavior, and hesitation has no such data apart from the entropy
split. Among them, only CLAS reports results on GSM8K, where it elicits step-by-step solutions from
prompts that ask only for the answer. Retraining each of them on the entropy split would change the
methods themselves. We therefore build the closest state-dependent baseline from our own pipeline.
The Euclidean probe of Appendix~\ref{app:euclid} is trained on the same entropy and lowers a
nonlinear readout of the hidden state, much as K-Steering and ODESteer follow the gradient of a
nonlinear classifier, and it differs from \HEST{} only in the curvature. CAA plays the same role for a fixed
direction, since we build it from the same entropy split.

\section{Hyperbolic Geometry and Proof of Proposition~\ref{prop:step}}
\label{app:geometry}

Throughout, $c>0$ is the curvature magnitude, so that the sectional curvature is $-c$, and
$n$ is the dimension of the ball.

\subsection{The Poincar\'e Ball}

The Poincar\'e ball model is
\begin{equation}
\Dball=\big\{z\in\mathbb{R}^{n}: c\lVert z\rVert^{2}<1\big\},
\qquad
g_{z}=(\lambda^{c}_{z})^{2}\,g^{E},
\qquad
\lambda^{c}_{z}=\frac{2}{1-c\lVert z\rVert^{2}},
\end{equation}
where $g^{E}$ is the Euclidean metric. The metric is a positive multiple of the Euclidean
one at every point, so the model is conformal: angles in $g$ and in $g^{E}$ agree. The
boundary sphere is at infinite hyperbolic distance from every interior point. In the
gyrovector formulation of \citet{ganea2018hyperbolic}, the M\"obius addition of $x,y\in\Dball$ is
\begin{equation}
x\Mob y=
\frac{\big(1+2c\langle x,y\rangle+c\lVert y\rVert^{2}\big)\,x+\big(1-c\lVert x\rVert^{2}\big)\,y}
{1+2c\langle x,y\rangle+c^{2}\lVert x\rVert^{2}\lVert y\rVert^{2}},
\label{eq:mobius}
\end{equation}
and the distance is
\begin{equation}
d_{c}(x,y)=\frac{2}{\sqrt{c}}\operatorname{artanh}\!\big(\sqrt{c}\,\lVert (-x)\Mob y\rVert\big).
\label{eq:dist}
\end{equation}
The exponential map at $x$ follows the geodesic that leaves $x$ with initial velocity
$v\in T_{x}\Dball$ for unit time. It is given by
\begin{equation}
\exp^{c}_{x}(v)=x\Mob\left(\tanh\!\Big(\frac{\sqrt{c}\,\lambda^{c}_{x}\lVert v\rVert}{2}\Big)\frac{v}{\sqrt{c}\,\lVert v\rVert}\right).
\label{eq:app-exp}
\end{equation}
At the origin $\lambda^{c}_{0}=2$, and Eq.~\ref{eq:app-exp} reduces to the map used in the
probe, Eq.~\ref{eq:probe}. For a smooth function $f$ on the ball, the Riemannian gradient is
$\gradR f(z)=(\lambda^{c}_{z})^{-2}\nabla f(z)$, and its length in the hyperbolic metric is
$\lVert\gradR f(z)\rVert_{g}=(\lambda^{c}_{z})^{-1}\lVert\nabla f(z)\rVert$.

\subsection{Proof of Part (i)}

Let $u$ have unit Euclidean norm and $w=\tanh(\sqrt{c}\,t/2)\,u/\sqrt{c}$, so that
$z'=z\Mob w$ and $\sqrt{c}\,\lVert w\rVert=\tanh(\sqrt{c}\,t/2)$. M\"obius addition satisfies
the left-cancellation law $(-z)\Mob(z\Mob w)=w$ \citep{ganea2018hyperbolic}. By Eq.~\ref{eq:dist},
\begin{equation}
d_{c}(z,z')=\frac{2}{\sqrt{c}}\operatorname{artanh}\!\big(\sqrt{c}\,\lVert w\rVert\big)
=\frac{2}{\sqrt{c}}\operatorname{artanh}\!\Big(\tanh\big(\tfrac{\sqrt{c}\,t}{2}\big)\Big)=t .
\end{equation}
Moreover, $z'=\exp^{c}_{z}(t\,v)$ with $v=u/\lambda^{c}_{z}$. This vector has unit length
in the hyperbolic metric, since $\lVert v\rVert_{g}=\lambda^{c}_{z}\lVert v\rVert=1$, and
substituting $t\,v$ into Eq.~\ref{eq:app-exp} gives $w$. \hfill$\square$

\subsection{Proof of Part (ii): the Radius Readout}

The gradient of $r(z)=\frac{2}{\sqrt{c}}\operatorname{artanh}(\sqrt{c}\lVert z\rVert)$ is
$\nabla r(z)=\lambda^{c}_{z}\,z/\lVert z\rVert$ for $z\neq0$, so $\lVert\gradR r\rVert_{g}=1$
and the steering direction is $u=-z/\lVert z\rVert$. Then $w$ is a negative multiple of $z$.
For collinear vectors $x=\rho_{1}\hat e$ and $y=\rho_{2}\hat e$ with $\lVert\hat e\rVert=1$,
Eq.~\ref{eq:mobius} simplifies: the numerator is $(\rho_{1}+\rho_{2})(1+c\rho_{1}\rho_{2})\,\hat e$ and the
denominator is $(1+c\rho_{1}\rho_{2})^{2}$, so $x\Mob y=\frac{\rho_{1}+\rho_{2}}{1+c\rho_{1}\rho_{2}}\,\hat e$. Writing
$\sqrt{c}\,\rho_{1}=\tanh\alpha$ and $\sqrt{c}\,\rho_{2}=\tanh\beta$ gives
$\sqrt{c}\,\frac{\rho_{1}+\rho_{2}}{1+c\rho_{1}\rho_{2}}=\tanh(\alpha+\beta)$. With
$\hat e=z/\lVert z\rVert$, $\alpha=\sqrt{c}\,r(z)/2$ and $\beta=-\sqrt{c}\,t/2$, the new point is
$z'=\tanh\big(\sqrt{c}\,(r(z)-t)/2\big)\,\hat e/\sqrt{c}$. For $0<t\le r(z)$, this gives
$r(z')=r(z)-t$. \hfill$\square$

\paragraph{The condition $t\le r(z)$ in our experiments.}
\label{app:radius}
Part (ii) requires the step to be no longer than the radius readout of the state. We evaluate
$r(z)$ at every hesitation point of the greedy test solutions, using the state that \HEST{} edits
when its gate fires. These states coincide with the edited states at the first hesitation point
of a solution and approximate them afterwards. Over the $117{,}032$ states of the seven radius
probes, the smallest value of $r(z)$ for each probe lies between $1.13$ and $1.55$ for
Qwen2.5-Math-1.5B-Instruct, whose largest step length is $t=1$. It lies between $0.88$ and $1.19$
for the other two models, which use $t=0.25$. None of these states therefore violates the
condition, and every radius step taken at them ends before the origin.

\subsection{Proof of Part (iii): the Busemann Readout}
\label{app:busemann}

A Busemann function is a renormalized distance to an ideal point on the boundary at infinity. For a
geodesic ray $\gamma$ with unit speed that ends at this point, it is
$B(y)=\lim_{s\to\infty}\big(d(y,\gamma(s))-s\big)$ \citep[Chapter II.8]{bridson2013metric}.
First take $c=1$. The function $B_{\xi}(y)=\log\big(\lVert\xi-y\rVert^{2}/(1-\lVert y\rVert^{2})\big)$
is the Busemann function of the unit Poincar\'e ball associated with the ideal point $\xi$,
normalized so that $B_{\xi}(0)=0$ \citep{ghadimi2021hyperbolic,bridson2013metric}. Its level sets
are horospheres, and a short calculation shows that the level set $\{B_{\xi}=\log k\}$ is the
Euclidean sphere with center $\xi/(1+k)$ and radius $k/(1+k)$, which is tangent to the
boundary at $\xi$.

The descent behavior of $B_{\xi}$ is easiest to see in the upper half-space model
$\mathbb{H}^{n}=\{x\in\mathbb{R}^{n}:x_{n}>0\}$ with metric $\lVert dx\rVert^{2}/x_{n}^{2}$.
A M\"obius transformation of $\mathbb{R}^{n}\cup\{\infty\}$ maps the unit ball isometrically onto
$\mathbb{H}^{n}$ and can be chosen to send $\xi$ to the point at infinity. Busemann functions are defined by distances
alone, so they are preserved by isometries, and the Busemann function of the point at
infinity is $B_{\infty}(x)=-\log x_{n}$ up to an additive constant, which follows directly from the
definition of Busemann functions \citep[Chapter II.8]{bridson2013metric}. Its
Euclidean gradient is $-e_{n}/x_{n}$, so its Riemannian gradient is $x_{n}^{2}(-e_{n}/x_{n})=-x_{n}e_{n}$,
which has length $1$ in the metric. The direction of steepest descent is therefore $+e_{n}$ at
every point, and the geodesic that leaves a point in this direction is the vertical line
through it. Along this line, hyperbolic arc length is $ds=dx_{n}/x_{n}$, so moving a
hyperbolic distance $\tau$ multiplies $x_{n}$ by $e^{\tau}$ and lowers $B_{\infty}$ by exactly $\tau$.
Transferring back to the ball, one step of hyperbolic length $\tau$ along the steepest-descent
geodesic of $B_{\xi}$ lowers $B_{\xi}$ by exactly $\tau$.

For general $c$, the dilation $z\mapsto\sqrt{c}\,z$ sends $\Dball$ onto the unit ball. Since
$\lambda^{1}_{\sqrt{c}z}=\lambda^{c}_{z}$, the metric of the unit ball pulls back to $c$ times the
metric of $\Dball$. The dilation therefore multiplies all distances by $\sqrt{c}$ and maps
geodesics to geodesics. For the readout $b(z)=B_{\xi}(\sqrt{c}\,z)$, it sends the steepest-descent geodesic of $b$ at $z$ to
that of $B_{\xi}$ at $\sqrt{c}\,z$. A step of length $t$ in $\Dball$ becomes a step of length
$\sqrt{c}\,t$ in the unit ball. By the previous paragraph, $b(z')=b(z)-\sqrt{c}\,t$.
\hfill$\square$

\subsection{Remarks on the Direction}
\label{app:direction}

\paragraph{The metric does not change the initial direction.}
Because the metric is conformal, the Riemannian gradient is a positive multiple of the
Euclidean gradient, and after normalization both give the same $u$. The geometry enters the method in
two other places. The step follows a geodesic of $\Dball$ and covers a fixed hyperbolic distance
(Proposition~\ref{prop:step}). The embedding itself is also shaped by the geometry, because the
probe is trained with the hyperbolic distance in Eq.~\ref{eq:loss} and with a readout defined on
the ball.

\paragraph{The Busemann direction depends on the state.}
The probe is defined in the coordinates of the ball, and the step is mapped back to the hidden
state from these coordinates. In the ball, the steepest-descent geodesics of $B_{\xi}$ are
the diameter through $\xi$ and circular arcs that meet the boundary orthogonally at $\xi$, so the direction $u$ differs from
state to state. This dependence is not an artifact of the coordinates. In the upper half-space
model, the descent directions look parallel, but the unit field $V=x_{n}\partial_{n}$ is not
parallel in the hyperbolic sense. For $n=2$, the Christoffel symbols of the metric give
$D_{\partial_{1}}V=-\partial_{1}\neq0$, where $D$ is the Levi-Civita connection. Parallel transport along a horosphere therefore
does not preserve the direction of steepest descent. The Euclidean counterpart of $b$, a linear
function, has a constant gradient, which is a parallel field.

\FloatBarrier
\section{Closed-Form Jacobian of the Probe}
\label{app:jacobian}

The update of Eq.~\ref{eq:pullback} needs $J=\partial E/\partial h$ at every iteration.
Because $E$ is a fixed composition, $J$ has a closed form. Write the forward pass of
Eq.~\ref{eq:probe} as
\begin{equation}
x=\frac{h-\mu}{\sigma},\quad
a=W_{1}x+b_{1},\quad
m=a\odot\Phi(a),\quad
q=\frac{W_{2}m+b_{2}}{\omega},\quad
z=\exp^{c}_{0}(q),
\end{equation}
where $\Phi$ is the standard normal distribution function, so that $m=\mathrm{GELU}(a)$ in its
exact form. With $\Phi'$ the standard normal density,
$\partial m_{k}/\partial a_{k}=\Phi(a_{k})+a_{k}\Phi'(a_{k})$ and
\begin{equation}
\frac{\partial q}{\partial h}=\omega^{-1}\,W_{2}\,\mathrm{diag}\big(\Phi(a)+a\odot\Phi'(a)\big)\,W_{1}\,\mathrm{diag}(1/\sigma).
\end{equation}
Let $\nu=\lVert q\rVert$ and $f(\nu)=\tanh(\sqrt{c}\,\nu)/(\sqrt{c}\,\nu)$, so that $z=f(\nu)\,q$.
Then
\begin{equation}
\frac{\partial z}{\partial q}=f(\nu)\,I_{n}+\frac{f'(\nu)}{\nu}\,qq^{\top},
\qquad
f'(\nu)=\frac{\operatorname{sech}^{2}(\sqrt{c}\,\nu)}{\nu}-\frac{\tanh(\sqrt{c}\,\nu)}{\sqrt{c}\,\nu^{2}},
\end{equation}
and $J=(\partial z/\partial q)(\partial q/\partial h)$, where $\omega$ is fixed after
training. The damping in Eq.~\ref{eq:pullback} is relative to the scale of the problem,
$\eta=10^{-3}\operatorname{tr}(JJ^{\top})/n$. It keeps the $n\times n$ solve stable when
$JJ^{\top}$ is ill-conditioned, for example near the boundary of the ball, where the radial
derivative $\operatorname{sech}^{2}(\sqrt{c}\,\nu)$ of the exponential map becomes much smaller
than its tangential derivative $f(\nu)$.

\paragraph{Accuracy of the mapping back.}
We reran the MATH-500 decodes of the development model with the Busemann readout at $t=0.5$ while logging
every steered token, and the rerun reproduces all $500$ solutions. Over the $4{,}426$ steered tokens
in $483$ solutions, the mean drop of the readout per solution lies between $0.4293$ and $0.4295$,
against $\sqrt{c}\,t=0.4294$ in Proposition~\ref{prop:step}. The mean hyperbolic distance between
$z$ and $E(h+\Delta h)$ lies between $0.5000$ and $0.5001$, against $t=0.5$. The edited states thus
realize the distance and the drop of the readout that Proposition~\ref{prop:step} prescribes to
within $10^{-4}$ on average in every solution. In the $56$ solutions with a single steered token,
the drop lies between $0.4293$ and $0.4294$ and the distance is $0.5000$, so the same holds for
individual edits. The Euclidean probes of Appendix~\ref{app:euclid} lower their readouts by the
prescribed amounts equally closely.

\FloatBarrier
\section{Probe Training Details}
\label{app:training}

\paragraph{Layer indexing.}
We index layers by the output of a decoder block, starting from zero, so our layer $\ell$ is
\texttt{hidden\_states[$\ell+1$]} in the Hugging Face convention. SEAL adds its vector to the input
of decoder layer $20$ of DeepSeek-R1-Distill-Qwen-1.5B and 7B, which is the output of our layer
$19$ \citep{chen2025seal}. CAA edits the output of the block with index $13$ in Llama 2 7B Chat
\citep{rimsky2024steering}. ODESteer reports layer $14$ of Llama-3.1-8B, and its code, which counts
layers from zero, uses index $13$ \citep{zhao2026odesteer}. The CAA and ODESteer settings both correspond to our layer $13$.

\paragraph{Why a middle layer.}
\HEST{} needs two things from the layer. The probe has to read the hesitation of the model from
the hidden state. In the early layers, the state mainly encodes the tokens of the input, and we
expect the uncertainty about the next token to be less readable there. The edit also needs enough
of the model after it. \HEST{} writes the edited state into the keys and values of all later
layers at that position (Section~\ref{sec:gate}), so that later tokens attend to it, whereas an
edit near the output mostly perturbs the logits of one token. Layer $19$ of the $28$ layers of the
Qwen models and layer $13$ of the $32$ layers of Llama-3.1-8B-Instruct lie between these two ends.
This agrees with the layer comparisons of prior work. SEAL reports that layers in the middle to
late part of the model perform best for steering \citep{chen2025seal}, and CAA sweeps over all
layers of Llama 2 7B Chat and finds that the steering effect peaks at layer $13$ and the adjacent
layers \citep{rimsky2024steering}. In the comparison of Appendix~\ref{app:ablation}, layer $19$
gains more than the earlier layer $9$ and the later layer $25$, and all three layers improve on
greedy decoding. Since the layer is taken from prior work, no test problem and no label enters its
choice.

\paragraph{Data.}
For each model and dataset, we decode $100$ training problems greedily once, with the prompt used
at evaluation. For every generated position, we record the layer-$\ell$ state and the entropy of
the next-token distribution. The state at the last prompt position emits the first generated
token, and we align states and entropies accordingly. The number of states ranges from
$15{,}821$ to $87{,}813$ per dataset (Table~\ref{tab:app-probe}).

\paragraph{Length limit of the training decodes.}
This step only collects training states, so its length limit plays a different role from the
limit of $3072$ new tokens used in all evaluations (Section~\ref{sec:setup}). The limit is
$1024$ new tokens for Qwen2.5-Math-1.5B-Instruct on MATH and GSM8K, $3072$ on OlympiadBench and
$2048$ for the other two models, which is at least twice the median length of the training
solutions in every setting. It cuts off $9$ of the $100$ solutions for Qwen2.5-Math-1.5B-Instruct
on MATH, $6$ on OlympiadBench and $5$ for Llama-3.1-8B-Instruct on MATH, and no solution in the
other settings. Because the model is causal, a cut leaves every recorded state and entropy
unchanged, and a longer limit would only add positions at the ends of these solutions. These ends
are nearly deterministic, since over the last $200$ tokens of the cut solutions the mean entropy is
at most $0.19$ nats and at most $6.5\%$ of the tokens exceed $1$ nat. A longer limit would therefore
add few hesitation points and mostly confident states.

\paragraph{Split, pairs and triplets.}
Twenty of the $100$ problems are held out for checkpoint selection. The split is by problem,
because states within one solution are strongly correlated. Ranking pairs $(i,j)$ are sampled
subject to $H_i<H_j-\delta$ with $\delta=0.3$ nats. Triplet anchors and positives come from
the same solution within five tokens of each other, and negatives come from a different
problem. Each optimization step uses $4096$ pairs and $4096$ triplets drawn from pools of
$400{,}000$ pairs and $200{,}000$ triplets.

\paragraph{Optimization.}
We use AdamW with learning rate $10^{-3}$ for the perceptron and $10^{-2}$ for the curvature
and the ideal point, weight decay $0.01$, $1500$ steps and margin $\gamma=0.1$. The output
dimension is $n=128$ and the hidden width is $1024$. We keep the checkpoint with the best pair
accuracy on the held-out problems. Table~\ref{tab:app-probe} reports the held-out quality of
the probes in one training run.

\begin{table}[!htb]
\caption{Held-out quality of the probes in one training run. States come from the greedy solutions of $100$ training problems, and $20$ of these problems are held out for checkpoint selection. Pair accuracy is the fraction of held-out pairs with an entropy gap above $0.3$ nats that the readout orders correctly, and it also selects the checkpoint. $\rho$ is the Spearman correlation between the readout and the entropy on held-out states, and $c$ is the learned curvature magnitude.}
\label{tab:app-probe}
\centering
\small
\setlength{\tabcolsep}{4pt}
\begin{tabular}{llrcccccc}
\toprule
& & & \multicolumn{3}{c}{Busemann} & \multicolumn{3}{c}{radius} \\
\cmidrule(lr){4-6}\cmidrule(lr){7-9}
Model & Probe data & States & Pair acc. & $\rho$ & $c$ & Pair acc. & $\rho$ & $c$ \\
\midrule
Qwen2.5-Math-1.5B & MATH & 50,717 & 0.880 & 0.554 & 0.74 & 0.842 & 0.472 & 0.53 \\
Qwen2.5-Math-1.5B & GSM8K & 27,212 & 0.855 & 0.501 & 0.71 & 0.829 & 0.429 & 0.40 \\
Qwen2.5-Math-1.5B & OlympiadBench & 87,813 & 0.900 & 0.547 & 0.78 & 0.876 & 0.478 & 0.38 \\
Llama-3.1-8B & MATH & 41,873 & 0.903 & 0.747 & 0.85 & 0.894 & 0.711 & 0.72 \\
Llama-3.1-8B & GSM8K & 15,821 & 0.871 & 0.706 & 0.91 & 0.868 & 0.698 & 0.73 \\
Qwen2.5-Math-7B & MATH & 54,889 & 0.893 & 0.620 & 0.70 & 0.869 & 0.540 & 0.46 \\
Qwen2.5-Math-7B & GSM8K & 28,462 & 0.853 & 0.444 & 0.78 & 0.824 & 0.380 & 0.58 \\
\bottomrule
\end{tabular}
\end{table}

\FloatBarrier
\section{Decoding Procedure}
\label{app:decoding}

Algorithm~\ref{alg:hest} states the decoding loop. The cache rollback makes the edit act on
the state whose next-token distribution is uncertain. On the greedy solutions of
Qwen2.5-Math-1.5B-Instruct, the median entropy at hesitation points is $1.23$ nats on
MATH-500 and $1.44$ nats on GSM8K, while one position later it is $0.075$ and $0.048$ nats.
Without the rollback, the edit would land on a state at which the model is already confident
and would leave the uncertain token unchanged.

\begin{algorithm}[h]
\caption{\HEST{} decoding for one sequence}
\label{alg:hest}
\begin{algorithmic}[1]
\STATE Run the prompt through the model; obtain the cache $C$ and the logits $o$ for the first token
\FOR{$i=1,\dots,T_{\max}$}
  \STATE $y_i\leftarrow\arg\max o$; append $y_i$ to the output; \textbf{if} $y_i$ is the end token \textbf{then} stop
  \STATE $(o, C)\leftarrow M(y_i\mid C)$ \COMMENT{ordinary forward pass}
  \IF{the entropy of $\mathrm{softmax}(o)$ exceeds $\theta$}
    \STATE Remove the last position from $C$
    \STATE $h\leftarrow$ layer-$\ell$ state of $y_i$; $z\leftarrow E(h)$; $z'\leftarrow$ Eq.~\ref{eq:geostep}
    \STATE $\Delta h\leftarrow$ Eq.~\ref{eq:pullback}, clipped to $\lVert\Delta h\rVert\le\lVert h\rVert$
    \STATE $(o, C)\leftarrow M(y_i\mid C)$ with $h\leftarrow h+\Delta h$ at layer $\ell$
  \ENDIF
\ENDFOR
\STATE \textbf{if} the output reached $T_{\max}=3072$ tokens without an end token, \textbf{return} the answer of greedy decoding
\end{algorithmic}
\end{algorithm}

The edit is applied once per hesitation point, but its effect persists. The edited state is
written into the keys and values of all later layers at that position, and every later token
attends to it. The procedure is causal and needs no information about the rest of the
generation. In a batch, the cache is rolled back for all sequences, and the edit is added only
to the rows whose entropy exceeds the threshold. For the other rows, the recomputed forward pass
is identical to the original one, so a hesitation point in any row costs one extra forward pass of
the whole batch. The greedy answer for the fallback is needed only for the decodes that reach the
limit.

\FloatBarrier
\section{Grading Protocol}
\label{app:judge}

Grading free-form mathematical answers needs two steps, extraction and equivalence. We use
one protocol for all models, because models differ in how they state their answers.

\paragraph{Extraction.}
We try four sources in order, namely the last \texttt{\textbackslash boxed\{\}} expression, an
explicit answer statement such as ``the answer is'', a restatement of the final answer and the
last number in the response. The model produces the restatement greedily, without steering and with
at most $64$ new tokens, when given its question, its reasoning and the prompt below.
\begin{promptbox}{Restatement prompt}
Generate the final answer for the query {query} based on the reasoning process {reasoning} in this format: "Answer: \[ \boxed{your_answer_here} \]". The entire answer should be contained completely within the \boxed{} command. Do not include any other text.
\end{promptbox}
The Qwen2.5-Math models box their answers in $97$ to $99\%$ of the solutions. For
Llama-3.1-8B-Instruct, the four sources are used in $31\%$, $6\%$, $59\%$ and $4\%$ of the
MATH-500 solutions and in $0\%$, $8\%$, $90\%$ and $2\%$ of the GSM8K solutions. A grader that
reads only \texttt{\textbackslash boxed\{\}} would therefore underestimate this model
substantially.

\paragraph{Equivalence.}
The extracted answer is compared with the reference in both raw and normalized form, using a
sequence of checks. Containers such as tuples, intervals and sets are compared element by
element. Answers are then compared by string match and numerically, with integers compared
exactly and decimals to a relative tolerance of $10^{-4}$. The remaining cases go to the
symbolic checker Math-Verify \citep{kydlicek2025mathverify} and, for scalars, to the
equivalence check of the Qwen2.5-Math evaluation toolkit \citep{yang2024qwen2}. Each check
runs under a time limit of $20$ seconds.

\paragraph{Grading errors and paired comparisons.}
All arms are graded by the same protocol, and the restatement is produced greedily by the unsteered
model, so the grade of a response depends only on its text. A solution whose text steering leaves
unchanged therefore receives the same grade as under greedy decoding. Grading errors can affect
the paired comparisons only through the changed solutions, since only a changed solution can be
counted as fixed or broken in Appendix~\ref{app:full}.

\FloatBarrier
\section{Selection of the Hyperparameters}
\subsection{Calibration of the Gate Threshold}
\label{app:theta}

The rule aims at the same number of steered tokens per solution as on the development model. Its
reference is the profile of Qwen2.5-Math-1.5B-Instruct at $\theta=1$, the percentage of its steered
decodes with $0$ to $10$, $11$ to $24$ and at least $25$ steered tokens, which is $72/18/10$ on
MATH-500 and $76/19/4$ on GSM8K. A new model has no steered decodes before $\theta$ is chosen.
For a candidate $\theta$, we therefore count the tokens with $H_i>\theta$ in each greedy solution
of the evaluation set, obtained by one teacher-forced pass, and compute the same profile. We take
the $\theta$ whose profile is closest to the reference in $L_1$ distance as the estimate
(Table~\ref{tab:app-theta}). On the development model, the greedy solutions give $73/19/9$ and
$76/20/5$, within an $L_1$ distance of $3$ of the reference, so the greedy count closely estimates
the number of steered tokens. At the selected thresholds, the other models steer a median of $4$ to
$6$ tokens per solution with the Busemann readout, against $5$ for the development model. The rule
uses no labels. It does read the unlabeled evaluation prompts and the model's own greedy solutions
to them, so the calibration adapts to the input distribution without seeing any answer. It matches counts per
solution rather than rates per token, because solution lengths differ across models and
datasets. For Llama-3.1-8B-Instruct, for example, the selected thresholds give $1.1\%$ of the
tokens on MATH-500 and $3.9\%$ on GSM8K, but a similar number of hesitation points per
solution.

\begin{table}[t]
\caption{Gate-threshold calibration. For each candidate $\theta$, the columns give the percentage of greedy tokens with $H>\theta$, the mean number of such tokens per solution, and the percentage of solutions with $0$--$10$, $11$--$24$ and at least $25$ of them. $L_1$ is the distance to the reference profile of Qwen2.5-Math-1.5B-Instruct at $\theta=1$, $72/18/10$ on MATH-500 and $76/19/4$ on GSM8K, which counts its steered tokens per solution (Appendix~\ref{app:theta}). The selected threshold (bold) minimizes $L_1$. Entropies come from one teacher-forced pass over each model's greedy solutions and use no labels. For Qwen2.5-Math-7B on GSM8K, the $9$ greedy solutions that reached the length limit are excluded.}
\label{tab:app-theta}
\centering
\footnotesize
\setlength{\tabcolsep}{3.5pt}
\begin{tabular}[t]{ccccc}
\toprule
$\theta$ & Tokens (\%) & Per solution & Profile (\%) & $L_1$ \\
\midrule
\multicolumn{5}{l}{\emph{Llama-3.1-8B, MATH-500}} \\
1.0 & 6.45 & 48.3 & 7/24/69 & 130 \\
1.4 & 3.31 & 24.8 & 33/29/39 & 79 \\
1.5 & 2.79 & 20.9 & 40/26/34 & 64 \\
1.6 & 2.34 & 17.5 & 47/25/28 & 50 \\
1.7 & 1.95 & 14.6 & 53/25/22 & 38 \\
1.8 & 1.62 & 12.1 & 60/24/16 & 24 \\
1.9 & 1.33 & 10.0 & 65/25/10 & 14 \\
\textbf{2.0} & 1.08 & 8.1 & 71/22/7 & 8 \\
2.2 & 0.69 & 5.2 & 83/14/3 & 22 \\
\midrule
\multicolumn{5}{l}{\emph{Llama-3.1-8B, GSM8K}} \\
1.0 & 10.03 & 19.0 & 25/54/21 & 103 \\
1.3 & 5.04 & 9.5 & 71/24/6 & 12 \\
\textbf{1.4} & 3.90 & 7.4 & 81/15/4 & 9 \\
1.5 & 3.00 & 5.7 & 89/9/2 & 25 \\
1.6 & 2.32 & 4.4 & 92/6/2 & 31 \\
1.7 & 1.72 & 3.3 & 94/4/2 & 35 \\
1.8 & 1.25 & 2.4 & 96/3/1 & 39 \\
2.0 & 0.65 & 1.2 & 98/1/1 & 43 \\
\bottomrule
\end{tabular}\hfill
\begin{tabular}[t]{ccccc}
\toprule
$\theta$ & Tokens (\%) & Per solution & Profile (\%) & $L_1$ \\
\midrule
\multicolumn{5}{l}{\emph{Qwen2.5-Math-7B, MATH-500}} \\
0.7 & 15.56 & 102.4 & 30/41/30 & 85 \\
0.8 & 14.12 & 93.0 & 52/25/22 & 39 \\
0.9 & 13.20 & 86.9 & 64/17/19 & 18 \\
\textbf{1.0} & 12.46 & 82.0 & 72/13/16 & 11 \\
1.1 & 11.87 & 78.2 & 78/8/14 & 20 \\
1.2 & 11.41 & 75.1 & 81/6/13 & 24 \\
1.3 & 11.02 & 72.6 & 83/5/12 & 26 \\
1.4 & 10.69 & 70.4 & 85/4/12 & 29 \\
1.5 & 10.42 & 68.6 & 86/3/11 & 30 \\
1.6 & 10.19 & 67.1 & 86/4/10 & 28 \\
1.8 & 9.83 & 64.7 & 87/5/8 & 30 \\
2.0 & 9.54 & 62.8 & 89/4/7 & 34 \\
\midrule
\multicolumn{5}{l}{\emph{Qwen2.5-Math-7B, GSM8K}} \\
0.7 & 3.94 & 12.0 & 59/35/6 & 35 \\
\textbf{0.8} & 2.66 & 8.1 & 81/16/3 & 9 \\
0.9 & 2.00 & 6.1 & 88/9/2 & 24 \\
1.0 & 1.57 & 4.8 & 93/6/1 & 33 \\
1.1 & 1.25 & 3.8 & 95/4/1 & 37 \\
1.2 & 1.01 & 3.1 & 96/3/1 & 39 \\
1.3 & 0.84 & 2.6 & 97/2/1 & 41 \\
1.4 & 0.72 & 2.2 & 98/1/1 & 43 \\
1.5 & 0.61 & 1.9 & 98/1/1 & 43 \\
1.6 & 0.54 & 1.6 & 98/1/1 & 43 \\
1.8 & 0.43 & 1.3 & 99/1/1 & 44 \\
2.0 & 0.37 & 1.1 & 99/0/1 & 45 \\
\bottomrule
\end{tabular}
\end{table}

\subsection{Selection of the Step Length}
\label{app:val}
The step length is selected on validation problems that are disjoint from the test sets and from
the $100$ training problems of the probe. For MATH, we start from $2{,}000$ problems of the training
split with $400$ problems per level, remove the $100$ problems of the probe and draw $500$ of the
remaining $1{,}900$ problems so that the number of problems at each level equals that of MATH-500
(Table~\ref{tab:app-val}). Subjects are not stratified. We checked by the text of the problems
that none of them occurs in MATH-500. For GSM8K, we remove the $100$ problems of the probe from the
$7{,}473$ training problems and draw $500$ uniformly. Their reference solutions have $3.23$ steps on
average, against $3.25$ for the test set, and $18.8\%$ of them have at least five steps, against
$17.1\%$. On each validation set, we run \HEST{} with one probe and
$t\in\{0.1,0.25,0.5,1\}$, and we keep the step length with the highest accuracy, taking the smaller
one in a tie (Table~\ref{tab:val}). The three probes of a setting use this step length on the test
set. The validation problems come from the training splits and are easier for the models than the
test problems, so their accuracies are not comparable with those of Table~\ref{tab:models}.
Appendix~\ref{app:mechanism} extends the sweep of the development model to $t=2$ and $t=5$.

\begin{table}[!htb]
\caption{The MATH validation set and MATH-500 by difficulty level (number of problems) and by
subject (\%).}
\label{tab:app-val}
\centering
\small
\begin{minipage}[t]{0.42\textwidth}
\centering
\begin{tabular}[t]{lcc}
\toprule
Level & Validation & MATH-500 \\
\midrule
1 & 43 & 43 \\
2 & 90 & 90 \\
3 & 105 & 105 \\
4 & 128 & 128 \\
5 & 134 & 134 \\
\bottomrule
\end{tabular}
\end{minipage}
\begin{minipage}[t]{0.54\textwidth}
\centering
\begin{tabular}[t]{lcc}
\toprule
Subject & Validation & MATH-500 \\
\midrule
Algebra & 23.8 & 24.8 \\
Intermediate algebra & 17.4 & 19.4 \\
Prealgebra & 16.6 & 16.4 \\
Geometry & 13.0 & 8.2 \\
Number theory & 10.2 & 12.4 \\
Precalculus & 10.2 & 11.2 \\
Counting and probability & 8.8 & 7.6 \\
\bottomrule
\end{tabular}
\end{minipage}
\end{table}

\FloatBarrier
\section{Additional Results}
\subsection{Results of the Individual Probes}
\label{app:full}
Table~\ref{tab:app-seeds} gives the accuracy of each of the three probes of a setting, with the
numbers of problems fixed and broken relative to greedy decoding. The probes differ only in the
random seed of their training, and they use the same $100$ training problems. The standard
deviation in Table~\ref{tab:models} is the sample standard deviation over the three probes. The
table also gives the number of problems for which the steered decode reached the length limit, so
that the fallback was used. The fallback matters most for Llama-3.1-8B-Instruct on MATH-500, where
greedy decoding itself reaches the length limit on $65$ of the $500$ problems.

\begin{table}[!htb]
\caption{\HEST{} with each of the three probes of a setting, at the selected step length $t$.
Accuracy in \%, with the numbers of problems fixed/broken relative to greedy decoding in parentheses, one
column per probe. Steered:
median percentage of steered tokens per solution. Fallbacks: number of steered decodes that reached the
length limit. The last two columns list the three probes in the same order.}
\label{tab:app-seeds}
\centering
\scriptsize
\setlength{\tabcolsep}{3pt}
\begin{tabular}{llccccccc}
\toprule
& & & & \multicolumn{3}{c}{\HEST{}, three probes} & & \\
\cmidrule(lr){5-7}
Model & Dataset & $t$ & Greedy & \multicolumn{3}{c}{Accuracy (fixed/broken)} & Steered & Fallbacks \\
\midrule
\multirow{2}{*}{Qwen2.5-Math-1.5B} & MATH-500 & 0.25 & 74.20 & 75.00 (9/5) & 74.60 (10/8) & 74.40 (9/8) & 1.05 / 1.10 / 1.15 & 6 / 2 / 4 \\
 & GSM8K & 1 & 84.84 & 85.29 (31/25) & 85.06 (25/22) & 84.69 (27/29) & 1.64 / 1.63 / 1.62 & 8 / 12 / 12 \\
\midrule
\multirow{2}{*}{Llama-3.1-8B} & MATH-500 & 0.5 & 50.80 & 52.00 (17/11) & 51.80 (17/12) & 51.80 (18/13) & 0.80 / 0.80 / 0.80 & 78 / 65 / 77 \\
 & GSM8K & 0.1 & 84.61 & 85.06 (17/11) & 84.23 (19/24) & 84.76 (14/12) & 3.07 / 3.00 / 2.94 & 7 / 6 / 5 \\
\midrule
\multirow{2}{*}{Qwen2.5-Math-7B} & MATH-500 & 1 & 84.00 & 84.60 (10/7) & 85.40 (13/6) & 85.20 (10/4) & 1.04 / 1.00 / 1.01 & 17 / 15 / 14 \\
 & GSM8K & 1 & 95.38 & 95.68 (9/5) & 95.38 (8/8) & 95.60 (10/7) & 2.02 / 2.01 / 2.01 & 9 / 8 / 8 \\
\bottomrule
\end{tabular}
\end{table}

\subsection{The Full MATH Test Set}
\label{app:math5000}
Table~\ref{tab:app-math5000} gives the numbers behind Figure~\ref{fig:math5000}. The run uses
Qwen2.5-Math-1.5B-Instruct with one probe, the threshold $\theta=1$ and the step length $t=0.25$
selected in Table~\ref{tab:val}, on all $5{,}000$ problems of the MATH test set.

\begin{table}[!htb]
\caption{\HEST{} on the full MATH test set, by difficulty level, by subject and by the number of
hesitation points in the greedy solution. Accuracy in \%, change in points. Problems: number of problems in the part. F/B: numbers of problems fixed and
broken.}
\label{tab:app-math5000}
\centering
\small
\begin{tabular}{llrcccc}
\toprule
& Part & Problems & Greedy & \HEST{} & Change & F/B \\
\midrule
\multirow{5}{*}{Level} & 1 & 437 & 94.74 & 96.34 & $+1.60$ & 8/1 \\
 & 2 & 894 & 89.04 & 89.82 & $+0.78$ & 11/4 \\
 & 3 & 1131 & 84.88 & 86.03 & $+1.15$ & 24/11 \\
 & 4 & 1214 & 73.89 & 74.63 & $+0.74$ & 32/23 \\
 & 5 & 1324 & 54.68 & 55.59 & $+0.91$ & 59/47 \\
\midrule
\multirow{7}{*}{Subject} & Algebra & 1187 & 91.66 & 92.25 & $+0.59$ & 15/8 \\
 & Prealgebra & 871 & 85.76 & 86.45 & $+0.69$ & 18/12 \\
 & Number theory & 540 & 79.26 & 80.37 & $+1.11$ & 15/9 \\
 & Counting and probability & 474 & 69.83 & 71.52 & $+1.69$ & 19/11 \\
 & Geometry & 479 & 64.51 & 65.97 & $+1.46$ & 13/6 \\
 & Precalculus & 546 & 62.45 & 64.10 & $+1.65$ & 18/9 \\
 & Intermediate algebra & 903 & 60.58 & 61.13 & $+0.55$ & 36/31 \\
\midrule
\multirow{3}{*}{Hesitation points} & 0--10 & 3640 & 89.73 & 90.25 & $+0.52$ & 48/29 \\
 & 11--24 & 910 & 47.14 & 49.45 & $+2.31$ & 57/36 \\
 & $\ge25$ & 450 & 21.33 & 23.11 & $+1.78$ & 29/21 \\
\midrule
\multicolumn{2}{l}{All problems} & 5000 & 75.82 & 76.78 & $+0.96$ & 134/86 \\
\bottomrule
\end{tabular}
\end{table}

\subsection{Hesitation Groups}
\label{app:groups}
Table~\ref{tab:app-groups} gives the problems that \HEST{} fixes and breaks in each group of
hesitation points, summed over the three probes of a setting. The groups are defined by the greedy
solution and were fixed before the analysis, so they are the same for every probe. In the group
with $0$ to $10$ hesitation points, which holds $71$ to $81\%$ of the problems of a setting, the
fixed and broken problems balance. The group with at least $25$ hesitation points has the largest
surplus of fixed problems. Table~\ref{tab:app-groups25} gives the accuracy of this group in each setting.

\begin{table}[!htb]
\caption{Numbers of problems fixed/broken by \HEST{} in each group of hesitation points of the greedy
solution, summed over the three probes of a setting.}
\label{tab:app-groups}
\centering
\small
\begin{tabular}{llcccc}
\toprule
Model & Dataset & 0--10 & 11--24 & $\ge25$ & All \\
\midrule
\multirow{2}{*}{Qwen2.5-Math-1.5B} & MATH-500 & 4/5 & 15/11 & 9/5 & 28/21 \\
 & GSM8K & 18/30 & 43/34 & 22/12 & 83/76 \\
\midrule
\multirow{2}{*}{Llama-3.1-8B} & MATH-500 & 26/20 & 21/16 & 5/0 & 52/36 \\
 & GSM8K & 20/19 & 18/17 & 12/11 & 50/47 \\
\midrule
\multirow{2}{*}{Qwen2.5-Math-7B} & MATH-500 & 5/3 & 5/11 & 23/3 & 33/17 \\
 & GSM8K & 12/7 & 5/11 & 10/2 & 27/20 \\
\midrule
\multicolumn{2}{l}{All settings} & 85/84 & 107/100 & 81/33 & 273/217 \\
\bottomrule
\end{tabular}
\end{table}

\begin{table}[!htb]
\caption{Accuracy (\%) in the group with at least $25$ hesitation points in the greedy solution.
Problems: number of problems in the group. \HEST{}: mean over the three probes. Change in points.}
\label{tab:app-groups25}
\centering
\small
\begin{tabular}{llcccc}
\toprule
Model & Dataset & Problems & Greedy & \HEST{} & Change \\
\midrule
\multirow{2}{*}{Qwen2.5-Math-1.5B} & MATH-500 & 43 & 18.60 & 21.71 & $+3.10$ \\
 & GSM8K & 64 & 43.75 & 48.96 & $+5.21$ \\
\midrule
\multirow{2}{*}{Llama-3.1-8B} & MATH-500 & 34 & 5.88 & 10.78 & $+4.90$ \\
 & GSM8K & 49 & 30.61 & 31.29 & $+0.68$ \\
\midrule
\multirow{2}{*}{Qwen2.5-Math-7B} & MATH-500 & 79 & 31.65 & 40.08 & $+8.44$ \\
 & GSM8K & 47 & 55.32 & 60.99 & $+5.67$ \\
\bottomrule
\end{tabular}
\end{table}

\FloatBarrier
\subsection{Significance Tests}
\label{app:sign}
Table~\ref{tab:app-sign} gives the exact two-sided sign test on the numbers of problems fixed and
broken, for the probe that selects the step length. The six tests on MATH-500 and GSM8K are not
significant, and the reason is the size of these test sets. \HEST{} changes the outcome of $1$ to
$6\%$ of the problems, so only $14$ to $56$ problems of a test set enter the test. On the full MATH
test set, $4.4\%$ of the problems change their outcome and $61\%$ of these are fixed. If the true
effect has this size, the exact test reaches $p<0.05$ with a probability of about $12\%$ on $500$
problems and about $34\%$ on $1{,}319$ problems, so a result that is not significant on a test set
of this size is expected even for a real effect. With $5{,}000$ problems, the test gives
$p=0.0015$, and on the $4{,}500$ problems outside MATH-500 it gives $p=0.0026$. In the group with
at least $25$ hesitation points, pooled over the six settings, the three probes fix and break
$28/8$, $28/14$ and $25/11$ problems, and the test gives $p=0.001$, $0.044$ and $0.029$.

\begin{table}[!htb]
\caption{Exact two-sided sign tests for \HEST{} with the probe that selects the step length.
Problems: number of problems in the test set. Fixed/broken: numbers of problems fixed and broken
relative to greedy decoding.}
\label{tab:app-sign}
\centering
\small
\begin{tabular}{llccc}
\toprule
Model & Dataset & Problems & Fixed/broken & $p$ \\
\midrule
\multirow{2}{*}{Qwen2.5-Math-1.5B} & MATH-500 & 500 & 9/5 & 0.42 \\
 & GSM8K & 1319 & 31/25 & 0.50 \\
\midrule
\multirow{2}{*}{Llama-3.1-8B} & MATH-500 & 500 & 17/11 & 0.34 \\
 & GSM8K & 1319 & 17/11 & 0.34 \\
\midrule
\multirow{2}{*}{Qwen2.5-Math-7B} & MATH-500 & 500 & 10/7 & 0.63 \\
 & GSM8K & 1319 & 9/5 & 0.42 \\
\midrule
\multicolumn{2}{l}{The six settings, pooled} & 5457 & 93/64 & 0.025 \\
Qwen2.5-Math-1.5B & Full MATH test set & 5000 & 134/86 & 0.0015 \\
\bottomrule
\end{tabular}
\end{table}

\FloatBarrier
\section{Baselines and Computational Cost}
\subsection{CAA}
\label{app:caa}

The CAA direction $v$ is the raw difference of means of Section~\ref{sec:setup}. Relative to
the median norm of the hidden states, its norm is $0.33$ and $0.24$ for
Qwen2.5-Math-1.5B-Instruct on MATH-500 and GSM8K, $0.30$ and $0.23$ for
Llama-3.1-8B-Instruct, and $0.18$ and $0.22$ for Qwen2.5-Math-7B-Instruct. A single \HEST{} edit
has a median relative norm of $0.01$ to $0.12$ at the selected step lengths
(Table~\ref{tab:app-controls}), and
it is applied to a small fraction of the tokens. Table~\ref{tab:app-caa} gives further
statistics. With the fallback of \HEST{}, CAA remains
below greedy decoding in all six settings.

\begin{table}[!htb]
\caption{CAA baseline in detail. CAA adds the raw difference-of-means vector (multiplier $1$) at every generated token, with the fallback of Section~\ref{sec:gate}. Cap: number of decodes that reached the 3072-token limit. Length: mean solution length relative to greedy decoding. Changed: percentage of solutions whose text differs from greedy decoding.}
\label{tab:app-caa}
\centering
\small
\setlength{\tabcolsep}{3pt}
\begin{tabular}{llccrcc}
\toprule
Model & Dataset & Greedy & CAA & Cap & Length & Changed (\%) \\
\midrule
Qwen2.5-Math-1.5B & MATH-500 & 74.20 & 72.40 ($-$1.80) & 36 & 1.26 & 100 \\
Qwen2.5-Math-1.5B & GSM8K & 84.84 & 84.38 ($-$0.45) & 8 & 0.91 & 99 \\
Llama-3.1-8B & MATH-500 & 50.80 & 44.20 ($-$6.60) & 131 & 1.33 & 100 \\
Llama-3.1-8B & GSM8K & 84.61 & 80.21 ($-$4.40) & 7 & 0.87 & 100 \\
Qwen2.5-Math-7B & MATH-500 & 84.00 & 83.20 ($-$0.80) & 10 & 0.99 & 97 \\
Qwen2.5-Math-7B & GSM8K & 95.38 & 95.00 ($-$0.38) & 6 & 0.95 & 98 \\
\bottomrule
\end{tabular}
\end{table}

\subsection{HARP and Computational Cost}
\label{app:harp}
HARP \citep{storai2025harp} treats tokens whose next-token entropy exceeds a threshold as points of
hesitation. At such a token it runs a second forward pass with dropout on the input embeddings and
mixes the logits of the two passes. We run the official implementation with its default
hyperparameters, an entropy threshold of $1.0$ bit, a dropout rate of $0.2$ and a mixing weight of
$0.5$, without tuning. HARP does not edit hidden states, and we give it the fallback of
Section~\ref{sec:gate}, as for every other method. Its result depends on the random
seed of the dropout, and Table~\ref{tab:app-harp} gives the three runs. HARP changes the outcome of many
more problems than \HEST{}: over the three runs it fixes $550$ problems and breaks $542$, whereas
\HEST{} fixes $273$ and breaks $217$ over its three probes. In the group with $0$ to $10$ hesitation
points, which greedy decoding mostly solves, HARP breaks more problems than it fixes in $13$ of the
$18$ runs.

\paragraph{Cost.}
We count token forward passes, where one pass takes one token position through the whole model
(Table~\ref{tab:app-cost}). \HEST{} removes one position from the cache at each hesitation point and
recomputes this position only, so it needs $1$ to $14\%$ more passes than greedy decoding with a
cache. When the greedy decode that the fallback uses is counted as well, the overhead is $4$ to
$45\%$, and it is largest for Llama-3.1-8B-Instruct on MATH-500, where the fallback is used most
often. HARP runs its second pass at $3.8$ to $18.8\%$ of the tokens. Its dropout acts on the
embeddings of the whole input, so this pass cannot reuse the cache and recomputes the whole
sequence. With a cache for the first pass, HARP needs $5.5$ to $30$ times the passes of greedy
decoding for the median problem, and more on average, because a few solutions run to the length
limit. The official implementation uses no cache at all, and against greedy decoding without a
cache it needs $3$ to $37\%$ more passes.

\begin{table}[!htb]
\caption{HARP in three runs per setting. Accuracy in \%, with the numbers of problems fixed/broken relative
to greedy decoding in parentheses, one column per run.}
\label{tab:app-harp}
\centering
\small
\setlength{\tabcolsep}{4pt}
\begin{tabular}{llccccc}
\toprule
& & & \multicolumn{3}{c}{HARP, three runs} & \\
\cmidrule(lr){4-6}
Model & Dataset & Greedy & \multicolumn{3}{c}{Accuracy (fixed/broken)} & Mean $\pm$ std \\
\midrule
\multirow{2}{*}{Qwen2.5-Math-1.5B} & MATH-500 & 74.20 & 73.60 (15/18) & 75.00 (21/17) & 73.80 (22/24) & $74.13\pm0.76$ \\
 & GSM8K & 84.84 & 84.61 (52/55) & 85.52 (63/54) & 84.69 (40/42) & $84.94\pm0.50$ \\
\midrule
\multirow{2}{*}{Llama-3.1-8B} & MATH-500 & 50.80 & 49.60 (42/48) & 49.40 (43/50) & 51.80 (47/42) & $50.27\pm1.33$ \\
 & GSM8K & 84.61 & 85.14 (57/50) & 83.24 (45/63) & 84.99 (55/50) & $84.46\pm1.05$ \\
\midrule
\multirow{2}{*}{Qwen2.5-Math-7B} & MATH-500 & 84.00 & 84.80 (9/5) & 85.40 (14/7) & 84.40 (11/9) & $84.87\pm0.50$ \\
 & GSM8K & 95.38 & 95.45 (5/4) & 95.68 (5/1) & 95.45 (4/3) & $95.53\pm0.13$ \\
\bottomrule
\end{tabular}
\end{table}

\begin{table}[!htb]
\caption{Cost in token forward passes, relative to greedy decoding with a cache. Greedy passes: number of
token forward passes of greedy decoding per problem. For HARP, the
mean over problems and, in parentheses, the median.}
\label{tab:app-cost}
\centering
\small
\begin{tabular}{llcccc}
\toprule
Model & Dataset & Greedy passes & \HEST{} & \HEST{} with fallback & HARP \\
\midrule
\multirow{2}{*}{Qwen2.5-Math-1.5B} & MATH-500 & 673 & $\times1.01$ & $\times1.04$ & $\times15$ ($\times7.8$) \\
 & GSM8K & 423 & $\times1.01$ & $\times1.06$ & $\times9.3$ ($\times6.6$) \\
\midrule
\multirow{2}{*}{Llama-3.1-8B} & MATH-500 & 859 & $\times1.07$ & $\times1.45$ & $\times42$ ($\times30$) \\
 & GSM8K & 289 & $\times1.03$ & $\times1.05$ & $\times25$ ($\times20$) \\
\midrule
\multirow{2}{*}{Qwen2.5-Math-7B} & MATH-500 & 762 & $\times1.14$ & $\times1.22$ & $\times216$ ($\times8.3$) \\
 & GSM8K & 418 & $\times1.06$ & $\times1.09$ & $\times71$ ($\times5.5$) \\
\bottomrule
\end{tabular}
\end{table}

\FloatBarrier
\section{Direction and Geometry Controls}
\subsection{Setup and Full Results}
\label{app:controls}
The controls of Table~\ref{tab:ablation} keep the layer, the threshold, the gate, the cache
recomputation and the fallback of \HEST{} and change only the edit. The \emph{fixed direction} is
the unit vector of the CAA direction of Section~\ref{sec:setup}, the mean of the low-entropy
states minus the mean of the high-entropy states of the same $100$ training problems. A
\emph{random direction} is a vector with independent standard normal entries in the hidden space,
normalized to unit length. One run uses the same vector at all problems and all hesitation points,
and the two datasets of a model use the same three vectors. At every hesitation point, the fixed
and the random directions are scaled to a fixed fraction of the norm of the hidden state. This
fraction is the median relative edit size $\lVert\Delta h\rVert/\lVert h\rVert$ of \HEST{} at the
selected step length, measured on the hesitation points of the probe-training solutions
(Table~\ref{tab:app-controls}). The \emph{reverse direction} is \HEST{} with $-t$ in place of $t$.
The \emph{Euclidean probe} is described in Appendix~\ref{app:euclid}. The fixed direction, the
reverse direction and the Euclidean probe are run once, with the probe that selects the step
length where a probe is needed.

\begin{table}[!htb]
\caption{Controls of Table~\ref{tab:ablation} in full. Accuracy in \%, with the numbers of problems
fixed/broken relative to greedy decoding in parentheses. Size: relative edit size of the fixed and
random directions. $t_{\mathrm{E}}$: step length of the Euclidean probe.}
\label{tab:app-controls}
\centering
\scriptsize
\setlength{\tabcolsep}{2.2pt}
\resizebox{\textwidth}{!}{
\begin{tabular}{llcccccccc}
\toprule
Model & Dataset & Size & $t_{\mathrm{E}}$ & Fixed & Euclidean & Reverse & Random 1 & Random 2 & Random 3 \\
\midrule
\multirow{2}{*}{Qwen2.5-Math-1.5B} & MATH-500 & 0.030 & 0.30 & 74.80 (10/7) & 73.80 (4/6) & 74.20 (8/8) & 74.00 (10/11) & 74.60 (8/6) & 74.00 (10/11) \\
 & GSM8K & 0.118 & 1.37 & 84.76 (24/25) & 84.69 (23/25) & 84.46 (24/29) & 83.85 (20/33) & 84.76 (19/20) & 85.06 (26/23) \\
\midrule
\multirow{2}{*}{Llama-3.1-8B} & MATH-500 & 0.058 & 0.55 & 49.40 (10/17) & 50.40 (15/17) & 50.20 (14/17) & 51.00 (14/13) & 50.40 (14/16) & 49.20 (10/18) \\
 & GSM8K & 0.011 & 0.06 & 84.23 (17/22) & 84.08 (10/17) & 84.31 (10/14) & 84.76 (17/15) & 84.31 (13/17) & 84.53 (18/19) \\
\midrule
\multirow{2}{*}{Qwen2.5-Math-7B} & MATH-500 & 0.098 & 1.26 & 83.80 (7/8) & 84.20 (11/10) & 82.20 (5/14) & 84.40 (8/6) & 84.20 (8/7) & 84.40 (7/5) \\
 & GSM8K & 0.108 & 1.15 & 95.53 (12/10) & 95.38 (10/10) & 95.30 (8/9) & 95.98 (11/3) & 95.45 (6/5) & 95.30 (5/6) \\
\bottomrule
\end{tabular}}
\end{table}

\subsection{Euclidean Probes}
\label{app:euclid}

The Euclidean probes of Section~\ref{sec:ablation} take the limit $c\to0$ of the probe of
Section~\ref{sec:method}. They keep the perceptron, the loss of Eq.~\ref{eq:loss}, the training
states, the held-out problems, the optimizer and the number of steps of
Appendix~\ref{app:training}. Only the curvature and the operations that depend on it change.

\paragraph{Formulas.}
As $c\to0$, the map $\exp^{c}_{0}$ tends to the identity, and the probe becomes
$E(h)=\phi\big((h-\mu)/\sigma\big)/\omega$. The conformal factor $\lambda^{c}_{z}$ tends to $2$, so
the metric tends to $4g^{E}$ and the distance to $d_{0}(z,z')=2\lVert z-z'\rVert$. The radius
readout tends to $r_{0}(z)=2\lVert z\rVert$, whose level sets are spheres around the origin. The
Busemann readout satisfies $b(z)/(2\sqrt{c})\to-\langle\xi,z\rangle$. We therefore use the linear
readout $b_{0}(z)=-\langle\xi,z\rangle$ with a learned unit vector $\xi$, which is half the Busemann
function of $d_{0}$ for the ideal point in the direction $\xi$. Its level sets are parallel
hyperplanes orthogonal to $\xi$, so its gradient is the same at every point. In Eq.~\ref{eq:geostep}, M\"obius addition tends to vector addition and
$\tanh(\sqrt{c}\,t/2)/\sqrt{c}$ tends to $t/2$, so the step becomes
\begin{equation}
z'=z+\frac{t}{2}\,u .
\label{eq:euclid-step}
\end{equation}
This step moves $z$ by $d_{0}(z,z')=t$, lowers $r_{0}$ by $t$ when $t\le r_{0}(z)$ and lowers
$b_{0}$ by $t/2$, which is the limit of Proposition~\ref{prop:step}. For the linear readout,
$u=\xi$ at every state, so the step in the probe space is the same vector everywhere, and only the
Jacobian of the probe makes the edit depend on the state. The Jacobian of
Appendix~\ref{app:jacobian} loses the factor of $\exp^{c}_{0}$. The mapping back to the hidden
state, the clipping and the gate are those of \HEST{}.

\paragraph{Step length.}
A step of length $t$ does not change the hidden state by the same amount in the two geometries,
since the probes differ in their metric and in their scale. We therefore match the size of the
edit instead of the step length. For every setting, we measure the median
$\lVert\Delta h\rVert/\lVert h\rVert$ of \HEST{} at its selected step length on the hesitation
points of the probe-training solutions, and we find by bisection the Euclidean step length
$t_{\mathrm{E}}$ that gives the same median (Table~\ref{tab:app-controls}). No test label enters
this choice.

\subsection{Diversity of the Steering Directions}
\label{app:field}

We apply the \HEST{} step to $400$ states of Qwen2.5-Math-1.5B-Instruct at layer $19$ whose
entropy exceeds $1$. The states come from $58$ MATH training problems, and we compare the
resulting edits $\Delta h$ (Table~\ref{tab:app-field}). A fixed steering vector would give pairwise cosines of
$1$ and a single principal component. The \HEST{} edits have a median pairwise cosine of $0.21$
for the Busemann readout, and $32$ principal components are needed to explain half of their
variance. Even the mean edit has a median cosine of only $0.47$ with the individual edits. The edits
are more similar within a problem than across problems, with pairwise cosines of $0.34$
within a problem and $0.21$ across problems for the Busemann readout.

\begin{table}[!htb]
\caption{Diversity of the \HEST{} edits on $400$ hesitation points of Qwen2.5-Math-1.5B-Instruct.
The columns give the median pairwise cosine between edits, with the 10th and 90th
percentiles in parentheses, the variance explained by the first principal component and by
the first five, the number of components needed for half of the variance, the median
relative size $\lVert\Delta h\rVert/\lVert h\rVert$, and the median cosine between each edit
and the mean edit.}
\label{tab:app-field}
\centering
\small
\begin{tabular}{lcccccc}
\toprule
Readout & Pairwise cos. & PC 1 & PCs 1--5 & PCs for 50\% & Size & Cos.\ to mean \\
\midrule
Busemann, $t=0.5$ & $0.210$ $(0.113,\,0.334)$ & $4.7\%$ & $17.6\%$ & $32$ & $0.060$ & $0.474$ \\
radius, $t=0.25$ & $0.251$ $(0.153,\,0.377)$ & $5.6\%$ & $18.8\%$ & $27$ & $0.017$ & $0.510$ \\
\bottomrule
\end{tabular}
\end{table}

\FloatBarrier
\section{What the Edit Changes at the Steered Token}
\label{app:mechanism}
Proposition~\ref{prop:step} describes the readout of the probe after the step. To see what the edit
does to the output of the model, we compare, at every steered position, the next-token
distribution before the edit with the distribution after it, for the same prefix
(Table~\ref{tab:app-mechanism}). \HEST{} acts through a small number of token choices. At the
selected step lengths, its edit changes the most likely token at $2.8$ to $10.7\%$ of the
hesitation points, more often than a random direction of the same size in all six settings, and
it lowers the entropy of the next-token distribution at $51$ to $64\%$ of them.

The comparison with the other directions shows that the gain of \HEST{} comes from which token it
selects, and not from a lower entropy alone. The fixed direction lowers the entropy most often, at
$54$ to $98\%$ of its edits, but it does not improve accuracy on average
(Table~\ref{tab:ablation}), whereas \HEST{} does.

These observations give the following account of how the gain arises. A hesitation point is a
close choice between candidate tokens. The model spreads its probability over several candidates,
greedy decoding takes the top one although its margin is small, and every later token of the
solution depends on this choice. In the example of Figure~\ref{fig:case}, the model hesitates on
which angle of the triangle is the right angle, greedy decoding writes the wrong one, and the final
answer is wrong. At such a point, \HEST{} moves the hidden state a small step along the geodesic
toward the ideal point of the probe, on whose side the confident states lie. The median edit is
$1$ to $12\%$ of the norm of the hidden state (Table~\ref{tab:app-controls}), so the most likely
token stays the same at $89$ to $97\%$ of the steered positions. \HEST{} therefore does not rewrite
the solution, and it changes the choice only where the candidates are close. One position later,
the model is usually confident again (Section~\ref{sec:gate}), which is why the entropy at the
edited position is not where the effect appears. The effect appears in the outcome of the
problems. Over the six settings, the forward step fixes $93$ problems and breaks $64$, the reverse
step with the same probe fixes $69$ and breaks $91$, and the random directions fix $224$ and break
$234$ (Table~\ref{tab:ablation}). Because the edit acts through these close choices, the gain is
concentrated on the solutions that contain many of them. A solution with at most $10$ hesitation
points receives few edits and is rarely changed, whereas in the group with at least $25$
hesitation points \HEST{} gains $0.68$ to $8.44$ points in the six settings
(Table~\ref{tab:app-groups25}).

Table~\ref{tab:app-sweep} varies the step length for the development model on the MATH validation
set, beyond the four values used for the selection. The effect of the edit grows smoothly with
the step. From $t=0.1$ to $t=5$, the edit changes the most likely token more often, from $2.5$ to
$39.1\%$ of the hesitation points, and it lowers the entropy more often, from $48.1$ to $61.0\%$.
Accuracy is highest at $t=0.25$ and remains stable beyond it. Even at $t=5$, where the edit is
almost half as large as the hidden state, \HEST{} solves only $14$ problems fewer than at the
selected step length, because it edits few positions.

\begin{table}[!htb]
\caption{Step length sweep of \HEST{} for Qwen2.5-Math-1.5B-Instruct on the $500$ MATH validation
problems. Correct: number of problems solved. Size: median relative edit size. The last two
columns are in \% of the edits.}
\label{tab:app-sweep}
\centering
\small
\begin{tabular}{cccccc}
\toprule
$t$ & Correct & Accuracy (\%) & Size & Most likely token changes & Entropy decreases \\
\midrule
0.1 & 413 & 82.6 & 0.013 & 2.5 & 48.1 \\
0.25 & \textbf{416} & \textbf{83.2} & 0.032 & 4.9 & 51.1 \\
0.5 & 414 & 82.8 & 0.064 & 7.8 & 52.4 \\
1 & 410 & 82.0 & 0.130 & 13.3 & 55.7 \\
2 & 406 & 81.2 & 0.267 & 25.7 & 59.5 \\
5 & 402 & 80.4 & 0.461 & 39.1 & 61.0 \\
\bottomrule
\end{tabular}
\end{table}

\begin{table}[!htb]
\caption{Effect of the edit on the next-token distribution at the steered positions of the test
decodes, in \% of the edits. Top: the most likely token changes. Bottom: the entropy decreases.
\HEST{}: mean over the three probes. Random: mean over the three random directions.}
\label{tab:app-mechanism}
\centering
\small
\begin{tabular}{llccccc}
\toprule
Model & Dataset & \HEST{} & Fixed & Euclidean & Reverse & Random \\
\midrule
\multicolumn{7}{l}{\emph{Most likely token changes (\%)}} \\
\multirow{2}{*}{Qwen2.5-Math-1.5B} & MATH-500 & 4.3 & 4.3 & 3.9 & 4.3 & 3.8 \\
 & GSM8K & 10.7 & 12.2 & 9.7 & 10.2 & 8.0 \\
\multirow{2}{*}{Llama-3.1-8B} & MATH-500 & 5.2 & 7.3 & 5.1 & 5.8 & 4.5 \\
 & GSM8K & 2.8 & 3.1 & 2.6 & 2.6 & 2.4 \\
\multirow{2}{*}{Qwen2.5-Math-7B} & MATH-500 & 3.3 & 3.2 & 2.9 & 2.5 & 2.0 \\
 & GSM8K & 8.2 & 10.6 & 5.6 & 6.1 & 5.4 \\
\midrule
\multicolumn{7}{l}{\emph{Entropy decreases (\%)}} \\
\multirow{2}{*}{Qwen2.5-Math-1.5B} & MATH-500 & 52.9 & 66.2 & 54.0 & 51.2 & 50.2 \\
 & GSM8K & 55.6 & 65.0 & 53.8 & 46.7 & 51.5 \\
\multirow{2}{*}{Llama-3.1-8B} & MATH-500 & 52.0 & 79.3 & 60.4 & 49.5 & 53.2 \\
 & GSM8K & 50.7 & 54.1 & 49.8 & 51.3 & 51.7 \\
\multirow{2}{*}{Qwen2.5-Math-7B} & MATH-500 & 64.3 & 97.6 & 65.8 & 28.7 & 51.5 \\
 & GSM8K & 52.1 & 65.6 & 49.4 & 49.4 & 55.8 \\
\bottomrule
\end{tabular}
\end{table}

\FloatBarrier
\section{Ablations at a Fixed Step Length}
\subsection{Positions, Threshold and Layer}
\label{app:ablation}

Table~\ref{tab:app-ablation} gives the full statistics of the ablations of positions, threshold
and layer in Section~\ref{sec:ablation}. They are run for the development model on MATH-500 with
one probe at the step length $t=0.5$, and every variant changes a single component at this step
length. All variants are graded with the protocol of
Appendix~\ref{app:judge}. The fallback of Section~\ref{sec:gate} changes the accuracy of every
variant of \HEST{} by at most $0.40$ points, so the conclusions do not depend on it.

\paragraph{Threshold and layer.}
The threshold $\theta=1$ was fixed before the ablation (Section~\ref{sec:calib}). Lowering it to
$\ln 2$ nearly triples the median fraction of steered tokens and lowers accuracy by $1.00$ point, and $\theta=0$
is the every-token variant. Raising it to $\ln 4$ or $\ln 8$ steers at most $0.5\%$ of the tokens,
and only $11$ and $4$ problems change their outcome. The probes for layers $9$ and $25$ are trained
with the same recipe on the same $100$ training problems. They steer the same fraction of tokens
as the probe for layer $19$ and give gains of $0.60$ and $0.20$ points, against $1.80$ points at
layer $19$.

\paragraph{Hesitation points and step boundaries.}
On the $500$ greedy MATH-500 solutions, a solution has a median of $6$ step boundaries and $5$
hesitation points. Only $356$ of the $3{,}974$ step starts are hesitation points ($9.0\%$), and
only $356$ of the $4{,}333$ hesitation points are step starts ($8.2\%$). A hesitation point lies
a median of $27$ tokens after the last step boundary. The two variants thus edit about as many
tokens, but at different positions.

\begin{table}[!htb]
\caption{Full statistics of the ablations on Qwen2.5-Math-1.5B-Instruct on MATH-500, Busemann readout,
$t=0.5$. Accuracy in \% with the change from greedy decoding in points, with the fallback and
without it. F/B: numbers of problems fixed and broken. Cap: number of steered decodes that reach the limit of $3072$
tokens. Steered: median and pooled fraction of steered tokens in \%. Edits: median number of
steered tokens per solution. Step: median over problems of the mean $\lVert\Delta h\rVert/\lVert h\rVert$ of a solution. For \HEST{}
it is measured at the hesitation points of the greedy solutions. Length: median number of tokens, against $470$ for greedy decoding.}
\label{tab:app-ablation}
\centering
\scriptsize
\setlength{\tabcolsep}{4pt}
\begin{tabular}{llcrrcccrr}
\toprule
& Variant & Accuracy & F/B & Cap & No fallback & Steered & Edits & Step & Length \\
\midrule
\multirow{3}{*}{Positions}
& Hesitation points (\HEST{}) & 76.00 (+1.80) & 16/7 & 6 & 76.00 & 1.1 / 1.6 & 5 & 0.064 & 472 \\
& Every token & 74.00 ($-$0.20) & 20/21 & 7 & 73.80 & 100 / 100 & 488 & 0.067 & 488 \\
& Step boundaries & 73.80 ($-$0.40) & 5/7 & 6 & 73.80 & 1.4 / 1.4 & 6 & 0.072 & 480 \\
\midrule
\multirow{4}{*}{Threshold}
& $\theta=\ln 2$ & 73.20 ($-$1.00) & 10/15 & 5 & 73.00 & 3.1 / 3.8 & 15 & 0.067 & 476 \\
& $\theta=1$ (\HEST{}) & 76.00 (+1.80) & 16/7 & 6 & 76.00 & 1.1 / 1.6 & 5 & 0.064 & 472 \\
& $\theta=\ln 4$ & 74.40 (+0.20) & 6/5 & 7 & 74.20 & 0.2 / 0.5 & 1 & 0.063 & 476 \\
& $\theta=\ln 8$ & 74.60 (+0.40) & 3/1 & 6 & 74.80 & 0.0 / 0.1 & 0 & 0.060 & 470 \\
\midrule
\multirow{3}{*}{Layer}
& $\ell=9$ & 74.80 (+0.60) & 10/7 & 3 & 74.80 & 1.1 / 1.6 & 5 & 0.052 & 474 \\
& $\ell=19$ (\HEST{}) & 76.00 (+1.80) & 16/7 & 6 & 76.00 & 1.1 / 1.6 & 5 & 0.064 & 472 \\
& $\ell=25$ & 74.40 (+0.20) & 12/11 & 6 & 74.00 & 1.1 / 1.6 & 5 & 0.069 & 470 \\
\bottomrule
\end{tabular}
\end{table}

\subsection{Radius Readout and OlympiadBench}
\label{app:fixedt}
Table~\ref{tab:qwen15} gives the test accuracy of the development model at four fixed step lengths
for both readouts, on MATH-500, GSM8K and OlympiadBench \citep{he2024olympiadbench}. These runs use
one probe per benchmark, and their step lengths are not selected on validation problems, so they
are not part of the protocol of Section~\ref{sec:setup}. From OlympiadBench, we take the $675$
English, text-only mathematics problems with open-ended answers, of which a fixed set of $100$
trains the probe and the other $575$ are used for evaluation.

\paragraph{Radius readout.}
The Busemann readout is the more robust of the two. It improves MATH-500 at every step length,
whereas the radius readout gains $1.00$ point at $t=0.25$ and depends more on the step length.
Together with the condition $t\le r(z)$ of Proposition~\ref{prop:step}, this is why the Busemann readout is
the default of \HEST{}.

\paragraph{OlympiadBench.}
OlympiadBench is a different regime from MATH-500 and GSM8K. The model hesitates on it
about twice as densely: $34\%$ of the problems have at least $25$ hesitation
points, against $9\%$ on MATH-500, and at $t=0.5$ \HEST{} steers $20$ tokens per solution on
average, against $9$. In this regime the radius readout is the better choice and gains up to $1.22$ points,
while the Busemann readout does not improve on greedy decoding there.

\begin{table}[!htb]
\caption{Accuracy (\%) of Qwen2.5-Math-1.5B-Instruct and change from greedy decoding (points) at
four step lengths $t$. Greedy decoding: $74.20$ on MATH-500, $84.84$ on GSM8K and $41.39$ on OlympiadBench. One probe per benchmark. Bold: best accuracy per benchmark.}
\label{tab:qwen15}
\centering
\footnotesize
\setlength{\tabcolsep}{3.2pt}
\begin{tabular}{lcccccc}
\toprule
& \multicolumn{3}{c}{\HEST{}, Busemann} & \multicolumn{3}{c}{\HEST{}, radius} \\
\cmidrule(lr){2-4}\cmidrule(lr){5-7}
$t$ & MATH-500 & GSM8K & OlympiadBench & MATH-500 & GSM8K & OlympiadBench \\
\midrule
0.1 & 74.80 (+0.60) & 84.53 ($-$0.30) & 41.39 (0.00) & 74.20 (0.00) & 84.53 ($-$0.30) & 41.04 ($-$0.35) \\
0.25 & 75.00 (+0.80) & 84.84 (0.00) & 41.39 (0.00) & 75.20 (+1.00) & 84.53 ($-$0.30) & 40.87 ($-$0.52) \\
0.5 & \textbf{76.00} (+1.80) & \textbf{85.37} (+0.53) & 40.00 ($-$1.39) & 74.40 (+0.20) & 84.38 ($-$0.45) & 42.26 (+0.87) \\
1 & 75.40 (+1.20) & 85.29 (+0.45) & 41.22 ($-$0.17) & 73.40 ($-$0.80) & 84.15 ($-$0.68) & \textbf{42.61} (+1.22) \\
\bottomrule
\end{tabular}
\end{table}

\FloatBarrier
\section{Probe States on Real Data}
\label{app:disks}

\begin{figure}[!htb]
\centering
\includegraphics[width=0.92\textwidth]{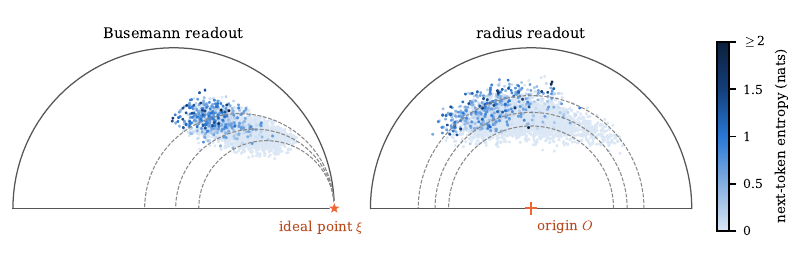}
\caption{Held-out states of Qwen2.5-Math-1.5B-Instruct on MATH-500, embedded by the Busemann probe
(left) and the radius probe (right). Each state $y=\sqrt{c}\,z$ is drawn at its norm and at its angle
to the ideal point $\xi$ (left) or to the first principal direction (right), which keeps its readout
exact. Rotations about the horizontal axis leave the readout unchanged, so a half disk shows all
states. Dashed curves are level sets of the readout at its $10$th, $50$th and $90$th percentiles.
Darker points have higher next-token entropy.}
\label{fig:app-disks-dev}
\end{figure}

\begin{figure}[!htb]
\centering
\includegraphics[width=\textwidth]{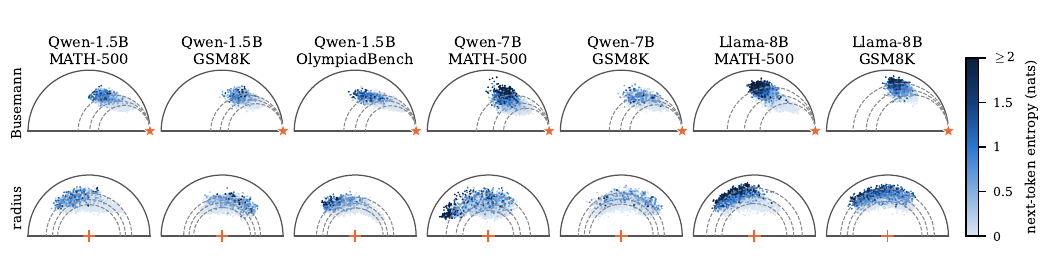}
\caption{Held-out states of every model and dataset, drawn as in
Figure~\ref{fig:app-disks-dev}. Top: Busemann probes. Bottom: radius probes.}
\label{fig:app-disks-all}
\end{figure}

Figure~\ref{fig:app-disks-dev} shows how the trained probes embed real hidden states of the
development model on MATH-500, and Figure~\ref{fig:app-disks-all} shows all models and datasets.
The states come from the $20$ validation problems that were held out from probe training, and each
panel shows $3{,}000$ of them sampled at random.

\paragraph{How the states are drawn.}
The probe space has $n=128$ dimensions, so any two-dimensional picture must drop information. We
keep the information that the readouts use. Let $y=\sqrt{c}\,z$, which lies in the unit ball. For
the Busemann probe, we draw $y$ at distance $\lVert y\rVert$ from the center and at its angle to
$\xi$, which lies between $0^{\circ}$ and $180^{\circ}$. The Busemann readout depends only on these
two quantities, because $\lVert\xi-y\rVert^{2}=1-2\langle\xi,y\rangle+\lVert y\rVert^{2}$. Each
state therefore keeps its exact readout, and the dashed level sets are exact horocycles. Rotating
a state about the axis through the origin and $\xi$ changes neither quantity, so all states related
by such a rotation are drawn at the same point, and a half disk is enough. A full disk would require
a second direction to decide on which side of the axis each state falls, and this choice would be
arbitrary. For the radius probe, the readout depends only on $\lVert y\rVert$, and we measure the
angle from the first principal direction of the states.

\paragraph{Comparison with the schematic.}
The real states follow the same ordering as the schematic in Figure~\ref{fig:pipeline}b,c, and the
pattern is similar across models and datasets. They differ from the schematic in four ways. The
states do not fill the ball, since $\lVert y\rVert$ lies between $0.39$ and $0.82$ for $90\%$ of
them, and almost no state lies at the origin or near the boundary. With the Busemann probe, confident and
hesitant states differ more in their angle to $\xi$ than in their norm. States with entropy
below $0.1$ nats have a median angle of $42^{\circ}$ to $66^{\circ}$, and states above the threshold
have a median angle of $66^{\circ}$ to $87^{\circ}$. Confident states are thus closer to $\xi$ in
direction rather than in norm. With the radius probe, confident states lie at the inner edge of the
occupied region rather than at the origin. Their median norm is $0.50$ to $0.62$, against $0.65$ to
$0.84$ for states above the threshold. Finally, the two groups overlap. Across the $14$ probes, the
Spearman correlation between the readout and the entropy of the plotted states ranges from $0.39$
to $0.76$.

These differences do not affect the method. \HEST{} uses only the gradient of the readout at the
current state and a fixed step length. Proposition~\ref{prop:step} holds wherever the state lies
in the ball, so a step of length $t$ lowers the readout by the same amount for every state. The
schematic is drawn to make the ordering of the readout easy to see, and Figure~\ref{fig:pipeline}
marks it as schematic.

\FloatBarrier
\section{The Example of Section~\ref{sec:example} in Full}
\label{app:case}

The problem is MATH-500 test problem 389 (geometry, level 3). Its diagram places the right
angle at $S$ and labels the side $RT$, the hypotenuse, with length $5$. The reference answer
is $\sqrt{21}/5$. We reproduce the problem and both solutions of Qwen2.5-Math-1.5B-Instruct
verbatim. The steered solution sets the right angle correctly at $S$, but it then calls $RS$
the hypotenuse and $RT$ a leg. The two labels are swapped consistently, so the final ratio,
and hence the answer, is correct even though these intermediate statements are not.

\outputfile{gray}{Problem}{tables/case389_problem.txt}
\outputfile{forkorange}{Greedy decoding: $189$ tokens, answer $1$, wrong}{tables/case389_greedy.txt}
\outputfile{hestblue}{\HEST{}, Busemann readout, $t=0.5$: $583$ tokens, $12$ steered, answer $\sqrt{21}/5$, correct}{tables/case389_hest.txt}

\end{document}